\documentclass[nonacm,sigconf]{acmart}

\renewcommand\footnotetextcopyrightpermission[1]{} % removes footnote with conference information in first column
\usepackage{float}
\usepackage{tcolorbox} 
\usepackage{listings}
\usepackage{subcaption}
\usepackage{tabularx} % Required for tabularx
\usepackage{graphicx} % Required for \resizebox 
\usepackage{amssymb}  % for up and down arrows
\usepackage{multirow}
\usepackage{mdwlist}
\usepackage{balance}

\tcbset{
    mybluebox/.style={
        colback=blue!5!white, % Light blue background
        colframe=blue!75!black, % Darker blue border
        boxrule=0.5mm, % Thickness of the border
        arc=4mm, % Rounded corners
        width=\textwidth, % Width of the box
        sharp corners,
        fonttitle=\bfseries,
    }
}

\AtBeginDocument{%
  }

\begin{document}

\title{E-SQL: Direct Schema Linking via Question Enrichment in
Text-to-SQL}


\author{Hasan Alp Caferoğlu}
\orcid{0009-0008-6687-9122}
\email{alp.caferoglu@bilkent.edu.tr}
\affiliation{%
  \institution{Bilkent University}
  \city{Ankara}
  \country{Turkey}
}

\author{Özgür Ulusoy}
\orcid{??????}
\email{oulusoy@cs.bilkent.edu.tr}
\affiliation{%
  \institution{Bilkent University}
  \city{Ankara}
  \country{Turkey}
}

\begin{abstract}
  Translating Natural Language Queries into Structured Query Language (Text-to-SQL or NLQ-to-SQL) is a critical task extensively studied by both the natural language processing and database communities, aimed at providing a natural language interface to databases (NLIDB) and lowering the barrier for non-experts. Despite recent advancements made through the use of Large Language Models (LLMs), significant challenges remain. These include handling complex database schemas, resolving ambiguity in user queries, and generating SQL queries with intricate structures that accurately reflect the user's intent. In this work, we introduce E-SQL, a novel pipeline specifically designed to address these challenges through direct schema linking and candidate predicate augmentation. E-SQL enhances the natural language query by incorporating relevant database items (i.e., tables, columns, and values) and conditions directly into the question and SQL construction plan, bridging the gap between the query and the database structure. The pipeline leverages candidate predicate augmentation to mitigate erroneous or incomplete predicates in generated SQLs.  Comprehensive evaluations on the BIRD benchmark illustrate that E-SQL achieves competitive performance, particularly excelling in complex queries with a 66.29\% execution accuracy on the test set. A further observation from our experiments reveals that incorporating schema filtering into the translation pipeline does not have a positive impact on performance when the most advanced proprietary LLMs are used. Additionally, our experiments with small LLMs highlight the importance and positive impact of enriched questions on their performance. Without fine-tuning, single-prompt SQL generation using enriched questions with DeepSeek Coder 7B Instruct 1.5v achieves 56.45\% execution accuracy on the BIRD development set.
\end{abstract}


%    <concept>
%        <concept_id>10002951.10002952.10003197.10010822.10010823</concept_id>
%        <concept_desc>Information systems~Structured Query Language</concept_desc>
%        <concept_significance>500</concept_significance>
%        </concept>
%    <concept>
%        <concept_id>10010147.10010178.10010179.10010180</concept_id>
%        <concept_desc>Computing methodologies~Machine translation</concept_desc>
%        <concept_significance>500</concept_significance>
%        </concept>
%  </ccs2012>

% \ccsdesc[500]{Information systems~Structured Query Language}
% \ccsdesc[500]{Computing methodologies~Machine translation}



\maketitle

\section{Introduction}
\label{sec:introduction}
The task of translating natural language queries into SQL (Text-to-SQL) has garnered considerable attention due to its potential to lower the technical barrier for non-experts and enhance the performance of querying or recommendation systems. Situated at the intersection of natural language processing (NLP) and database management, this task aims to enable users to interact with databases through simple, natural language queries, without requiring extensive knowledge of SQL syntax or database schema structures. Despite advancements in utilizing large language models (LLMs) for Text-to-SQL, a significant performance gap of around 20\% still remains between the best-performing models and human-level accuracy, underscoring that even the most sophisticated pipelines are not yet suitable for real-world deployment as a natural language interface to databases~\cite{li-2024-bird-sql}.

Prior to the emerge of LLMs~\cite{brown2020gpt-3, ouyang2022instructgpt, openai2023gpt4, touvron2023llama, touvron2023llama2, meta2024llama3, chen2021codex, chowdhery2022palm, anil2023palm2, li2023starcoder}, a wide range of studies ~\cite{zhong2017seq2sql, guo2019irnet, lin2020bridge, fu2023catsql, choi2021ryansql, wang2020ratsql, usta-2021-DBTagger, usta-2023-xDBTagger, lei-etal-2020-re-examining-schema-linking-slsql} focused on building encoder-decoder based neural network architectures utilizing recurrent neural networks ~\cite{hochreiter1997lstm, cho-etal-2014-gru-gated-recurrent-unit} and various pre-trained language models~\cite{devlin2019bert, raffel2020t5, yu2021grappa, clark2020electra}. These early approaches established a foundation but were often limited in handling complex queries or schemas.

LLMs have shown substantial potential in the Text-to-SQL task, yielding impressive results across various benchmarks. To further enhance the reasoning capabilities of LLMs, a variety of in-context learning (ICL) techniques have been introduced, including chain-of-thought (CoT) prompting~\cite{wei2022COT}, question decomposition~\cite{zhou2023leasttomost, khot-2023-decomposed-prompting}, self-consistency~\cite{wang2023selfconsistency}, and others ~\cite{xi-2023-self-polish, huang2023selfimprove, madaan-2023-self-refine, zhang2023-automatic-cot}. Although many of these strategies have been successfully applied to Text-to-SQL translation pipelines~\cite{pourreza-2023-dinsql, Gao-2024-dail-sql, qu-etal-2024-ta-sql, li-2024-codes, talaei2024chess}, improving LLM reasoning specifically from the perspective of question refinement remains relatively underexplored. 

Beyond in-context learning, LLM performance can also be improved through fine-tuning or training from scratch. However, these techniques are resource-intensive, requiring significant computational resources and large volumes of task-specific annotated data. While proprietary models are less frequently fine-tuned for Text-to-SQL~\cite{maamari2024deathschemalinkingtexttosql-distyl-ai}, promising results have been achieved through the fine-tuning of numerous open-source models~\cite{Pourreza-2024-dts-sql, Gao-2024-dail-sql, pourreza-2023-dinsql, li-2024-codes, talaei2024chess, wang-2024-macsql}.

\begin{figure*}
  \centering
  \includegraphics[width=\linewidth]{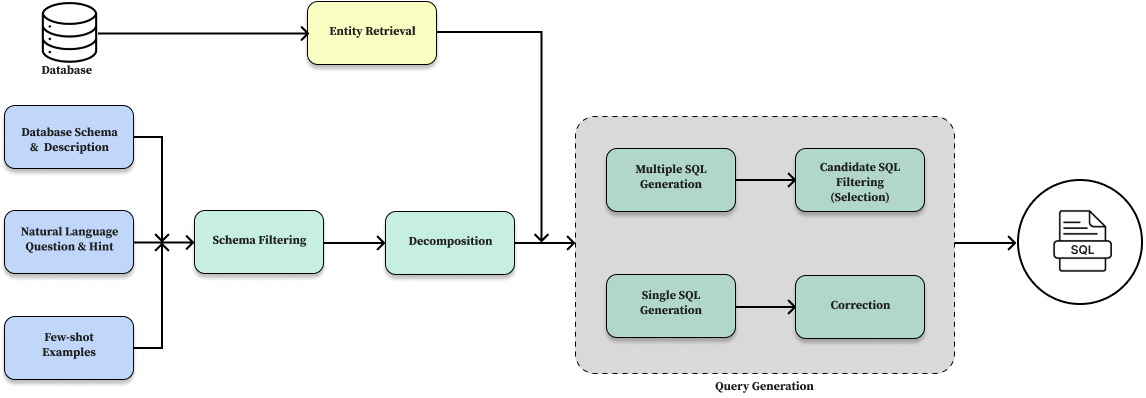}
  \caption{Overview of the general pipeline for the Text-to-SQL translation task, highlighting the key modules: Schema Filtering, Question Decomposition, Entity Retrieval, and Query Generation. The modular design allows for variation in the usage of these components, depending on the preferred pipeline configuration.}
  \label{fig:general-text2sql-pipeline}
\end{figure*}

\autoref{fig:general-text2sql-pipeline} demonstrates the general pipeline and modules used in prior works. A critical component of the Text-to-SQL task is schema linking, which involves connecting the sense of natural language query to the database schema. Although various methods have been proposed to enhance schema linking, it remains a core challenge. Schema filtering, a technique commonly used to eliminate irrelevant database items, has been widely adopted to reduce noise for downstream tasks. While both neural network-based~\cite{Li-2023-Resdsql, li-2024-codes} and LLM-based~\cite{dong-2023-c3, talaei2024chess, lee-2024-mcs-sql, pourreza-2023-dinsql, qu-etal-2024-ta-sql} schema filtering techniques have been explored, our findings align with those of Maamari et al.~\cite{maamari2024deathschemalinkingtexttosql-distyl-ai}, indicating that schema filtering can result in performance degradation when the latest generation LLMs are employed. Additionally, several studies~\cite{Gao-2024-dail-sql, pourreza-2023-dinsql, talaei2024chess, qu-etal-2024-ta-sql, wang-2024-macsql, Pourreza-2024-dts-sql, maamari2024deathschemalinkingtexttosql-distyl-ai} reveal that providing database-related information in response to a query significantly enhances performance.

In this work, we introduce \textit{E-SQL:  Direct Schema Linking via Question Enrichment in Text-to-SQL}\footnote{The complete code required to reproduce the reported results is publicly available on our GitHub repository\url{https://anonymous.4open.science/r/E-SQL_Direct_Schema_Linking}}, a novel pipeline designed to directly address the schema linking challenge through question enrichment and candidate predicate augmentation. We explore the improvement of both LLM reasoning and schema linking from the perspective of question enrichment. E-SQL enhances the natural language query representation by incorporating relevant database elements—such as tables, columns, and values—directly into the query, along with an associated SQL construction plan. This approach is augmented by generating candidate predicates, which reduce the likelihood of erroneous or incomplete SQL predicates. This methodology differs from traditional schema filtering techniques, which have been commonly used to simplify the schema presented to the model. The E-SQL pipeline consists of four main modules: Candidate SQL Generation (CSG), Candidate Predicate Generation (CPG), Question Enrichment (QE), and SQL Refinement (SR).

In the Candidate SQL Generation (CSG) module, an initial SQL query is generated. This query is then parsed to extract values and operations from its predicates. The Candidate Predicate Generation (CPG) module uses these extracted elements to find similar values from the database and constructs candidate predicates. Using the candidate predicates, the Question Enrichment (QE) module instructs the LLM to incorporate relevant database items and possible predicates into the natural language question, while concurrently formulating SQL construction steps as its reasoning process. These steps are then utilized to produce a fully enriched query. Simultaneously, the candidate SQL query is executed to identify potential execution errors. Finally, in the SQL Refinement (SR) module, the candidate SQL query is either refined or a new SQL query is generated, utilizing the enriched question, candidate predicates, and any identified execution errors.

The impact of schema filtering, a widely adopted technique in previous research, is also explored on our pipeline. We incorporate an additional schema filtering module into our pipeline, where the LLM is instructed to select only the database tables and columns relevant to the query while eliminating others. Following this, the Filtered Schema Correction technique is applied to resolve any inconsistencies between the filtered schema and the original database schema. Our experiments demonstrate that schema filtering can negatively affect performance when applied in conjunction with the most advanced proprietary LLMs. Instead, direct schema linking through question enrichment and candidate predicate augmentation provides a more reliable strategy for accurate SQL generation, particularly in complex cases.

Through an ablation study, we illustrate the effectiveness of each module within the pipeline for the Text-to-SQL task. In particular, our question enrichment module significantly improves performance on challenging questions, yielding nearly a 5\% increase in accuracy.

We evaluate E-SQL on the Spider~\cite{yu-etal-2018-spider} and BIRD~\cite{li-2024-bird-sql} benchmarks, well-known standard datasets for the Text-to-SQL task, and demonstrate its ability to handle complex queries while maintaining competitive performance with state-of-the-art methods. Our findings suggest that the integration of enriched questions, SQL generation steps, and candidate predicates leads to more accurate SQL generation, particularly for complex queries involving multiple conditions and joins. Therefore, our approach establishes a new paradigm for schema linking and prompt augmentation by leveraging question enrichment and candidate predicate augmentation in the context of Text-to-SQL translation. 

The key contributions of our work can be summarized as follows:

\begin{itemize}
\item We propose a new paradigm for schema linking through question enrichment, which leads to direct schema linking by incorporating related database items and potential conditions into the natural language question. Fully enriched queries further guide LLMs in SQL construction by providing explicit logical steps.  
\item To the best of our knowledge, we are the first to enhance both LLM reasoning capabilities and schema linking performance through the use of a question enrichment module, which composes database integrated questions, in Text-to-SQL translation task.
\item We propose a candidate predicate generation technique leveraging the LIKE operator and demonstrate its positive impact by augmenting prompts with candidate predicates.
\item  Our experiments also confirm the potential drawbacks of traditional schema filtering techniques when integrated into a Text-to-SQL translation pipeline like ours, which leverages the most advanced proprietary LLMs.
\item  We demonstrate the importance and positive impact of database-integrated questions, including logical SQL construction steps, on the performance of small LLMs in the task of Text-to-SQL translation, achieved without requiring fine-tuning.
\end{itemize}

\begin{figure*}
  \centering
  \includegraphics[width=\linewidth]{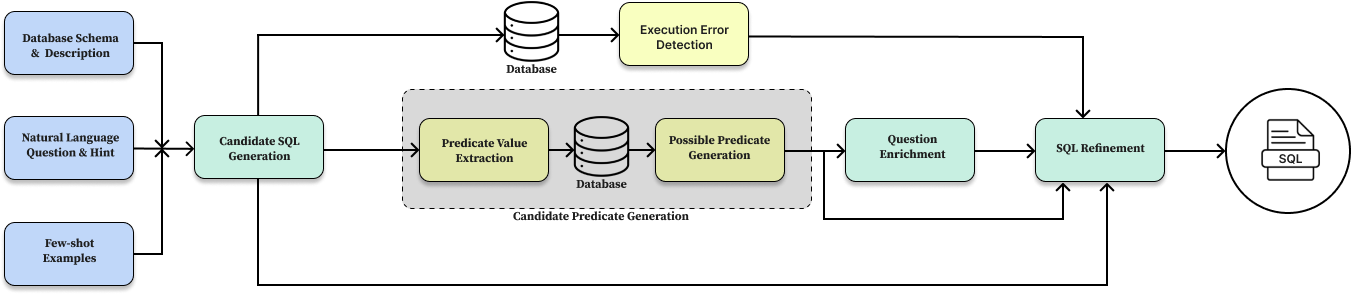}
  \caption{Overview of the proposed E-SQL pipeline with candidate predicate generation, question enrichment, SQL refinement modules, and without schema filtering module.}
  \label{fig:our-pipeline}
\end{figure*}

\section{Related Work}

Before the emergence of LLMs, supervised fine-tuning approaches in Text-to-SQL translation focused on encoder - decoder architectures that utilized recurrent neural networks (RNNs) ~\cite{zhong2017seq2sql, guo2019irnet, usta-2021-DBTagger}, pre-trained language models (PLMs) ~\cite{lin2020bridge, fu2023catsql, lei-etal-2020-re-examining-schema-linking-slsql}, convolutional neural networks (CNNs) ~\cite{choi2021ryansql} and graph neural networks (GNNs)~\cite{wang2020ratsql}. These methods encoded natural language questions alongside database schema to establish schema linking and generated SQL queries through sequence generation ~\cite{zhong2017seq2sql, lin2020bridge}, grammar-based methods ~\cite{guo2019irnet, choi2021ryansql, wang2020ratsql}, or sketch-based slot-filling strategies. These approaches provided a foundational understanding for Text-to-SQL tasks, paving the way for more advanced solutions.

The emergence of both proprietary and open-source LLMs \cite{brown2020gpt-3, ouyang2022instructgpt, touvron2023llama, chen2021codex, hui2024qwen2, deepseek-coder} has marked a significant shift in the field. Thanks to their advanced reasoning and comprehension capabilities, the research community has increasingly focused on harnessing the power of these models for Text-to-SQL tasks. 

\subsection{LLM Reasoning}

Advanced reasoning techniques are crucial for improving LLM performance on complex tasks. While prompt design is important, methods that enhance intrinsic reasoning, such as breaking down problems or refining formulations, have led to significant advancements in LLM capabilities.

The Chain-of-Thought (CoT) approach~\cite{wei2022COT} significantly improves the performance for multi-step reasoning tasks by guiding LLMs to generate intermediate reasoning steps. Kojima et al.~\cite{kojima-2022-LLMS-are-zero-shot-reasoners} explored a simple yet effective prompting, "Let's think step by step", to uncover the reasoning ability of the language models in zero-shot mode. Decomposing complex questions into simpler sub-questions has been explored~\cite{zhou2023leasttomost, khot-2023-decomposed-prompting, dua-etal-2022-successive-prompting-decomposing-questions}. Techniques like Self Consistency~\cite{wang2023selfconsistency} apply majority voting to select consistent answers, while Self-Improve~\cite{huang2023selfimprove} and Self-Refine~\cite{madaan-2023-self-refine} iteratively refine responses through self-generated data and feedback, respectively. Moving beyond answer generation, Xi et al.~\cite{xi-2023-self-polish} demonstrated significant gains by refining problem contexts and questions through the Self-Polish technique.

To the best of our knowledge, the refinement and generation of high-quality user queries, particularly with embedded reasoning that facilitates the construction of correct SQL queries, has not been explored in the literature for the Text-to-SQL translation task. Our work addresses this gap by proposing a methodology that produces clear, schema-aware queries enriched with reasoning elements, explicitly guiding the generation of accurate SQL queries and overcoming schema linking challenges.

\subsection{Schema Linking and Filtering}
Translating a natural language query into SQL requires a clear understanding of both the language used in the question and the structure of the database. This process is facilitated by schema linking, which bridges the gap between the query and the database schema, ensuring that words or phrases in the query are accurately matched to the relevant database elements, such as tables, columns, or values. 

Previous works have shown that removing irrelevant database elements can improve schema linking and enhance model performance in Text-to-SQL tasks by reducing the likelihood of schema-based hallucinations. This process, known as schema filtering or schema pruning, has been extensively studied. RESDSQL~\cite{Li-2023-Resdsql} and CodeS~\cite{li-2024-codes}  address this by classifying schema items based on their relevance to the natural language query and then filtering them according to their classification probabilities after ranking. While DIN-SQL~\cite{pourreza-2023-dinsql} uses a single step to select only the question relevant database tables and columns, C3~\cite{dong-2023-c3}, CHESS~\cite{talaei2024chess}, and MCS-SQL~\cite{lee-2024-mcs-sql} first filter database tables (table linking) and then select the most appropriate columns (column linking) from the previously filtered tables, achieving better schema filtering. For schema linking and filtering, the TASL module of TA-SQL~\cite{qu-etal-2024-ta-sql} generates a list of schema entities (tables and columns) from an LLM-generated dummy SQL query, which is then used to create symbolic representations. Our approach also leverages the initially generated candidate SQL query; however, rather than using it for schema filtering, we extract possible database values from conditions to generate potential predicates, which are then used in downstream tasks as part of data augmentation, explained in Section \ref{subsec:cpg}. Gao et al.~\cite{Gao-2024-dail-sql} explored the impact of different schema representations and demonstrated that representing the database schema as code, instead of natural language representation, leads to a better understanding of the question-to-SQL mapping, thereby improving schema linking. While previous works have demonstrated the positive impact of schema filtering on schema linking, Maamari et al.~\cite{maamari2024deathschemalinkingtexttosql-distyl-ai} argue that explicit schema filtering remains useful for less capable large language models (LLMs), but it becomes unnecessary with the latest, more advanced LLMs such as GPT-4o. Our experiments with single-step schema filtering implemented in the E-SQL pipeline corroborate this finding, indicating that explicit schema filtering can be redundant and can result in performance degradation when applied within pipelines leveraging advanced LLMs such as GPT-4o and GPT-4o-mini ~\cite{openai2023gpt4}.

\subsection{Data Augmentation}
One essential aspect of data augmentation involves providing relevant content to the LLM through prompts. Augmenting prompts with items pertinent to the query is crucial for improving Text-to-SQL translation performance. Depending on the sub-task, prompts are typically enriched with explanations of database items, database values selected based on similarity, sub-task specific examples, and database schema—represented either as code or natural language—that are filtered or unfiltered, along with decomposed queries and candidate SQL queries~\cite{Gao-2024-dail-sql, pourreza-2023-dinsql, talaei2024chess, qu-etal-2024-ta-sql, wang-2024-macsql, Pourreza-2024-dts-sql, maamari2024deathschemalinkingtexttosql-distyl-ai}. In our work, we incorporate  database item explanations and database values similar to the query in prompt. Additionally, we include execution errors for candidate queries to provide valuable feedback during query refinement. Unlike prior work, we propose a novel approach to further reduce predicate errors and enhance LLM performance by introducing candidate conditions with exact database values, effectively bridging the gap between the natural language query and the database.

\subsection{SQL Query Generation and Correction} 
The final step in Text-to-SQL tasks is generating an SQL query that accurately answers the user's natural language question. While some approaches, like C3~\cite{dong-2023-c3}, employ zero-shot generation, few-shot prompting techniques are more commonly used to enhance performance. Methods such as self-consistency~\cite{wang2023selfconsistency} and multi-choice selection are employed by models like C3, DAIL-SQL, CHESS, and MSC-SQL~\cite{dong-2023-c3, Gao-2024-dail-sql, talaei2024chess, lee-2024-mcs-sql, maamari2024deathschemalinkingtexttosql-distyl-ai}. However, these techniques lead to high computational costs due to large number of generation steps for a single query. To address missing or redundant keywords in generated SQLs, Pourreza and Rafiei~\cite{pourreza-2023-dinsql} introduced a self-correction module. With MAC-SQL, Wang et al.~\cite{wang-2024-macsql} proposed a refiner agent that evaluates the syntactic correctness and executability of the generated SQL, ensuring non-empty result generation. In our work, we propose a SQL refinement module similar to MAC-SQL~\cite{wang-2024-macsql}, where we refine the SQL after enriching the question and augmenting the prompt with candidate predicates.

\section{Methodology}
\label{sec:methodology}

Our work focuses on bridging the gap between the database schema and users' natural language queries by enriching the questions with database items through keyword mappings and expanding them with SQL generation steps and reasoning. Our approach can be characterized as direct schema linking via query enrichment with relevant database elements. To further minimize errors such as incorrect table, column, or missing value usage in predicates, we augment the prompt with all possible predicates after extracting them.

Our proposed method consists of four main modules, as illustrated in \autoref{fig:our-pipeline}: Candidate SQL Generation (CSG), Candidate Predicate Generation (CPG), Question Enrichment (QE), and SQL Refinement (SR). Additionally, recognizing the mixed outcomes of database schema filtering observed in prior work, we include a Schema Filtering and Filtered Schema Correction module (SF). However, our experiments on our pipeline reveal that schema filtering can lead to performance degradation when integrated into Text-to-SQL pipelines with the most advanced large language models (LLMs), corroborating the findings of Maamari et al.~\cite{maamari2024deathschemalinkingtexttosql-distyl-ai}. Detailed explanations of each module are provided in the following subsections.

\subsection{Database Description and Value Selection}
\label{subsec:db-description-sample-selection}
In real-world databases, description files serve as a valuable resource, offering detailed information about database items. Large-scale databases often contain numerous description files with content that can be extensive. Another crucial piece of information for LLMs is the actual data values within the database. However, augmenting the prompt with all available descriptions and database values is impractical due to the context window limitations of LLMs and the associated computational costs. Another effective perspective to prompt augmentation involves carefully selecting relevant database-related information while filtering out content unrelated to the query, thereby minimizing noise and enhancing model performance. 

Li et al.~\cite{Li-2023-Graphix-T5}, Scholak et al.~\cite{scholak-etal-2021-picard}, and Li et al.~\cite{Li-2023-Resdsql} utilize the Longest Common Substring (LCS) algorithm to identify database values related to the query. However, the time complexity of the LCS algorithm can significantly slow down the retrieval process. To address this issue, Li et al.~\cite{li-2024-codes} propose a "coarse-to-fine" matching framework, which initially extracts hundreds of potentially relevant values from the entire database using BM25, followed by the application of the LCS algorithm to retrieve the most query-relevant values. Talaei et al.~\cite{talaei2024chess} employ a different approach by first extracting keywords from the natural language query with the assistance of an LLM, and then retrieving similar values from the database using a Locality Sensitive Hashing (LSH) based hierarchical retrieval strategy. DIN-SQL~\cite{pourreza-2023-dinsql} takes a more focused approach by extracting possible cell values directly from the query with the help of an LLM, similar to the keyword extraction phase of CHESS~\cite{talaei2024chess}. In our work, we employ the BM25 ranking algorithm not only to retrieve the most relevant database values but also to identify the most pertinent database descriptions. We augment prompts with the 10 most relevant database values for each column and the 20 most relevant database description sentences, as determined by the BM25 ranking algorithm. Additionally, when providing database values for each column, we ensure that any columns containing \texttt{"NULL"} values also include \texttt{"NULL"} as part of the selected values. This guarantees that the LLM is aware of potential null entries, allowing it to incorporate conditions such as \texttt{"IS NOT NULL"} when needed.

\subsection{Candidate SQL Generation Module (CSG)}
\label{subsec:csg}
When provided with appropriate information, LLMs can effectively establish strong connections between the database schema and the natural language query, resulting in the generation of mostly accurate SQL queries. Although these SQL queries may occasionally contain errors, such as incorrect table or column usage or missing values in predicates, they typically avoid incorporating completely irrelevant database items. To leverage this capability, we extract potentially useful information from the initially generated SQL queries in subsequent steps to enhance data quality and, consequently, improve model performance. Qu et al.~\cite{qu-etal-2024-ta-sql} generate and employ dummy SQL for extracting schema entities. Further details on how dummy SQL is utilized in our approach are provided in Section \ref{subsec:cpg}.

Our candidate SQL generation module incorporates three randomly selected samples, each from a different database than the one associated with the current query, across various difficulty levels from the few-shot data. These samples are ordered to present simpler question-SQL pairs first, followed by more challenging pairs. After providing database schema code representation as suggested in DAIL-SQL~\cite{Gao-2024-dail-sql}, the prompt is further augmented with selected database descriptions and relevant database values for each column as described in the previous subsection \ref{subsec:db-description-sample-selection}. To enhance the LLM's reasoning capabilities, we use the phrase "Let's think step by step" and instruct the LLM to generate reasoning steps, as proposed by Kojima et al.~\cite{kojima-2022-LLMS-are-zero-shot-reasoners} and Wei et al.~\cite{wei2022COT}.

\subsection{Candidate Predicate Generation (CPG)}
\label{subsec:cpg}
Determining the correct predicates to use in the SQL query is a crucial step in Text-to-SQL translation. Successfully establishing the relationship between database items and the query is essential. However, even the most advanced LLMs sometimes struggle to generate accurate predicates. For a predicate in an LLM-generated SQL query, assuming the correct operation is used, there are six possible cases:

\begin{enumerate}
    \item The predicate is correct syntactically, schematically, and semantically. In other words, executing the \texttt{"SELECT * FROM <table> WHERE <table>.<column> <operation> <value>"} query would yield results without any execution errors or an empty set.
    \item The table and column are correct, but the value used in the predicate is incomplete or contains extraneous characters or words. As shown in \autoref{fig:wrong-predicate-case-2}, the generated predicate uses \texttt{"Fresno"} instead of \texttt{"Fresno County Office of Education"} as a value, leading to an incorrect SQL despite the absence of execution errors.
    \item The table and value are correct, but the wrong column is selected in the predicate. In other words, another column in the selected table contains the value. \autoref{fig:wrong-predicate-case-3} shows the wrong column usage while the selected table and value are correct. 
    \item The correct table is selected, but both the column and the value are incorrect or contain missing or extraneous parts. As shown in \autoref{fig:wrong-predicate-case-4}, \texttt{"T1.'County Name' = 'Fresno'"} is generated as part of the predicate, whereas it should be  \texttt{"T1.'District Name' = 'Fresno County Office of Education'"}.
    \item The value is generated correctly, but the wrong table and column are selected, meaning the generated value should be used but belongs to another table and one of its columns.
    \item The table, column, and value are wrong and totally irrelevant to the question.
\end{enumerate}

To address the possible errors outlined in items (2) to (5), we propose the Candidate Predicate Generation (CPG) module, as seen in the \autoref{fig:our-pipeline}, to enhance downstream tasks by augmenting prompts, enabling the LLM to be aware of possible predicates and generate correct ones. In the CPG module, we first parse the candidate SQL query to extract the values and operations used in predicates. We then utilize the \texttt{LIKE} operator to retrieve all potential values from the database by executing the following query:

\begin{quote}
\begin{verbatim}
SELECT DISTINCT <COLUMN> 
FROM <TABLE> 
WHERE <COLUMN> LIKE '%<VALUE>%';
\end{verbatim}
\end{quote}

Here, \texttt{<VALUE>} represents a token extracted from the values in candidate SQL query. This process results in a list of possible predicates, formatted as "\texttt{<table>.<column> <operation> <value>}," which is used in downstream tasks.

\begin{figure}[htbp]
\centering
\begin{tcolorbox}[colback=orange!10!white, colframe=orange!10!white, width=\columnwidth]
\lstset{
    breaklines=true,
    basicstyle=\ttfamily\scriptsize,
    columns=fullflexible,
    frame=none,
    backgroundcolor=\color{orange!10!white}, % Light orange background
    xleftmargin=0pt,
    xrightmargin=0pt,
    showspaces=false,
    showstringspaces=false,
    keepspaces=true, % Keep spaces exactly as they are, without any additional indentation
    breakindent=0pt, % No indent for wrapped lines
    aboveskip=0pt,
    belowskip=0pt,
    morekeywords={Question, Evidence, Predicted, SQL, Gold}, % Keywords to highlight
    keywordstyle=\bfseries % Makes the keywords bold
}

\begin{lstlisting}
Question: 
Please list the zip code of all the charter schools in Fresno County Office of Education.

Predicted SQL: 
SELECT T2.Zip 
FROM frpm AS T1 INNER JOIN schools AS T2 ON T1.CDSCode = T2.CDSCode 
WHERE T1.`District Name` = 'Fresno' AND T1.`Charter School (Y/N)` = 1

Gold SQL:
SELECT T2.Zip 
FROM frpm AS T1 INNER JOIN schools AS T2 ON T1.CDSCode = T2.CDSCode 
WHERE T1.`District Name` = 'Fresno County Office of Education' AND T1.`Charter School (Y/N)` = 1
\end{lstlisting}
\end{tcolorbox}
\caption{Example for the generation of incomplete value in the predicate explained in Section \ref{subsec:cpg}, case (2).}
\label{fig:wrong-predicate-case-2}
\end{figure}

\begin{figure}[htbp]
\centering
\begin{tcolorbox}[colback=orange!10!white, colframe=orange!10!white, width=\columnwidth]
\lstset{
    breaklines=true,
    basicstyle=\ttfamily\scriptsize,
    columns=fullflexible,
    frame=none,
    backgroundcolor=\color{orange!10!white}, % Light orange background
    xleftmargin=0pt,
    xrightmargin=0pt,
    showspaces=false,
    showstringspaces=false,
    keepspaces=true, % Keep spaces exactly as they are, without any additional indentation
    breakindent=0pt, % No indent for wrapped lines
    aboveskip=0pt,
    belowskip=0pt,
    morekeywords={Question, Evidence, Predicted, SQL, Gold}, % Keywords to highlight
    keywordstyle=\bfseries % Makes the keywords bold
}

\begin{lstlisting}
Question: 
Please list the zip code of all the charter schools in Fresno County Office of Education.

Predicted SQL: 
SELECT T2.Zip 
FROM frpm AS T1 INNER JOIN schools AS T2 ON T1.CDSCode = T2.CDSCode 
WHERE T1.`County Name` = 'Fresno County Office of Education' AND T1.`Charter School (Y/N)` = 1

Gold SQL:
SELECT T2.Zip 
FROM frpm AS T1 INNER JOIN schools AS T2 ON T1.CDSCode = T2.CDSCode 
WHERE T1.`District Name` = 'Fresno County Office of Education' AND T1.`Charter School (Y/N)` = 1
\end{lstlisting}
\end{tcolorbox}
\caption{Example for correct table and value but wrong column in the predicate explained the Section \ref{subsec:cpg}, case (3).}
\label{fig:wrong-predicate-case-3}
\end{figure}

\begin{figure}[htbp]
\centering
\begin{tcolorbox}[colback=orange!10!white, colframe=orange!10!white, width=\columnwidth]
\lstset{
    breaklines=true,
    basicstyle=\ttfamily\scriptsize,
    columns=fullflexible,
    frame=none,
    backgroundcolor=\color{orange!10!white}, % Light orange background
    xleftmargin=0pt,
    xrightmargin=0pt,
    showspaces=false,
    showstringspaces=false,
    keepspaces=true, % Keep spaces exactly as they are, without any additional indentation
    breakindent=0pt, % No indent for wrapped lines
    aboveskip=0pt,
    belowskip=0pt,
    morekeywords={Question, Evidence, Predicted, SQL, Gold}, % Keywords to highlight
    keywordstyle=\bfseries % Makes the keywords bold
}

\begin{lstlisting}
Question: 
Please list the zip code of all the charter schools in Fresno County Office of Education.

Predicted SQL: 
SELECT T2.Zip 
FROM frpm AS T1 INNER JOIN schools AS T2 ON T1.CDSCode = T2.CDSCode 
WHERE T1.`County Name` = 'Fresno' AND T1.`Charter School (Y/N)` = 1

Gold SQL:
SELECT T2.Zip 
FROM frpm AS T1 INNER JOIN schools AS T2 ON T1.CDSCode = T2.CDSCode 
WHERE T1.`District Name` = 'Fresno County Office of Education' AND T1.`Charter School (Y/N)` = 1
\end{lstlisting}
\end{tcolorbox}
\caption{Example for correct table but wrong column and value selection in the predicate explained in Section \ref{subsec:cpg}, case (4).}
\label{fig:wrong-predicate-case-4}
\end{figure}

\subsection{Schema Filtering and Filtered Schema Correction Module (SF)}
\label{subsec:sf}

In prior works, schema filtering—eliminating database tables and columns irrelevant to the query—has been shown to improve model performance by reducing the likelihood of generating irrelevant schema items in SQL queries. One approach involves instructing the LLM to first select the relevant database tables and then choose the most relevant columns from those tables ~\cite{dong-2023-c3, talaei2024chess, lee-2024-mcs-sql}. Another approach filters the database schema by considering both tables and columns simultaneously in a single step ~\cite{pourreza-2023-dinsql}. Additionally, some methods leverage dummy SQL queries to extract relevant database entities for schema filtering ~\cite{qu-etal-2024-ta-sql}. Li et al. propose CODES ~\cite{li-2024-codes} and train a schema classifier to predict relevance scores following RESDSQL ~\cite{Li-2023-Resdsql}. In our work, we adopt a single-step schema filtering approach, extracting only the relevant tables and their associated columns.

However, we observed that the filtered schema may not always be compatible with the original schema, where a selected column might belong to a different table according to the original database schema. When such issues arise, they can negatively impact the SQL generation process and lead to a decline in performance. To address this, we propose a filtered schema correction strategy, which checks for mismatches between the filtered schema and the original schema, subsequently correcting them accordingly. While previous research has demonstrated the benefits of schema filtering for schema linking and overall Text-to-SQL translation performance, our experiments show that incorporation of schema filtering into our pipeline result in performance decrease when used with the latest proprietary LLMs, a finding that aligns with the work of Maamari et al. ~\cite{maamari2024deathschemalinkingtexttosql-distyl-ai}. Detailed experimental results are provided in Section \ref{sec:sf-experiment-discussion}. Consequently, we have chosen not to include a schema filtering module in our E-SQL pipeline.

\subsection{Question Enrichment Module (QE)}
\label{subsec:qe}
\begin{figure}[htbp]
  \centering
  \begin{subfigure}{\linewidth}
    \centering
    \includegraphics[width=\linewidth]{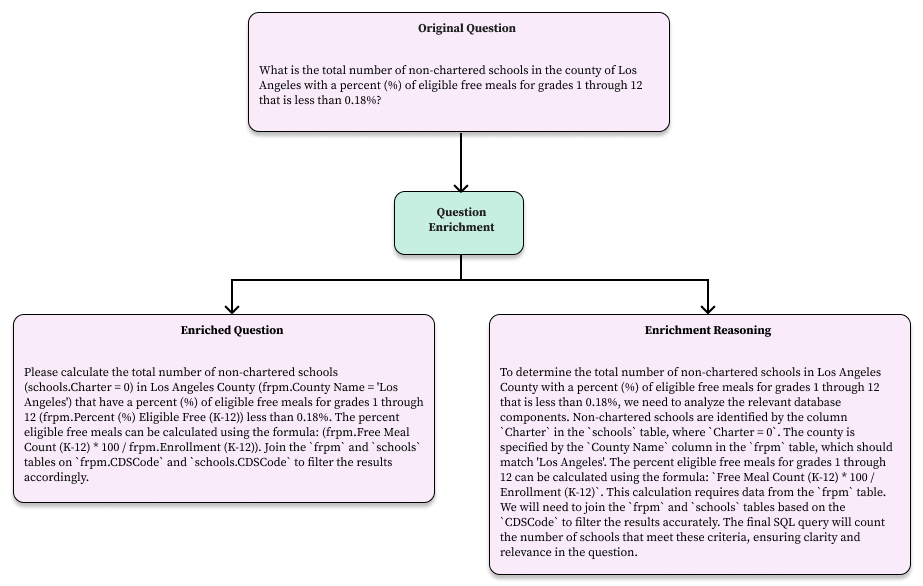}
    \caption{Question enrichment example}
    \label{fig:question-enrichment-example}
  \end{subfigure}
  
  \vspace{3em} % Adds some space between the two figures
  
  \begin{subfigure}{\linewidth}
    \centering
    \includegraphics[width=\linewidth]{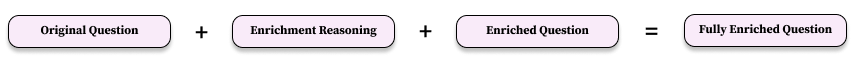}
    \caption{Concatenation of original question, enrichment reasoning, and enriched question.}
    \label{fig:fully-enriched-question-logic}
  \end{subfigure}
  
  \caption{(a) Question enrichment example and (b) Fully enriched question construction.}
  \label{fig:QE-module}
\end{figure}

To enhance schema linking, various LLM reasoning improvement techniques, such as chain-of-thought (CoT) ~\cite{wei2022COT}, question decomposition ~\cite{zhou2023leasttomost, khot-2023-decomposed-prompting, dua-etal-2022-successive-prompting-decomposing-questions}, and self-consistency ~\cite{wang2023selfconsistency} have been applied to Text-to-SQL translation tasks. Almost all prior works leveraging LLMs utilize chain-of-thought reasoning. Question decomposition has been applied to Text-to-SQL tasks by Pourreza et al. ~\cite{pourreza-2023-dinsql} and Wang et al. \cite{wang-2024-macsql}. Self-consistency and multi-choice selection techniques have been employed by models like C3\cite{dong-2023-c3}, DAIL-SQL ~\cite{Gao-2024-dail-sql}, CHESS ~\cite{talaei2024chess}, and MSC-SQL ~\cite{lee-2024-mcs-sql}. Another key approach is schema filtering, which eliminates irrelevant database items and augments the prompt with database values related to the query, thereby narrowing the gap between the query and the database schema ~\cite{dong-2023-c3, talaei2024chess, lee-2024-mcs-sql, qu-etal-2024-ta-sql}. Previous paradigms have largely overlooked enhancing the reasoning and schema linking capabilities of LLMs through question reformulation. Focusing on this aspect, we propose a novel question enrichment strategy that directly links natural language questions to the database schema by expanding the question with relevant database items and incorporating logical steps to generate accurate SQL queries as shown in \autoref{fig:QE-module}. 

In the Question Enrichment module (QE), we instruct the LLM to refine the original question by incorporating relevant database items (tables, columns, and values) and conditions. This process makes the question more understandable, coherent, well-aligned with the database schema. Additionally, an SQL construction plan, generated during the question-database integration as part of the reasoning process, is appended to the enriched question. Together, this plan and the enriched question form the fully enriched question which explicitly incorporates the necessary SQL components and logical steps, guiding the LLM to generate accurate SQL queries. The creation process of a fully enriched question, which combines the original question, the enriched question, and the enrichment reasoning, is illustrated in ~\autoref{fig:fully-enriched-question-logic}.  A few-shot strategy is applied to generate refined questions, using randomly sampled question-SQL pairs from the development set that have been manually annotated and made publicly available. We annotate 12 questions for each difficulty level: simple, moderate, and challenging. Since the manually annotated questions are taken from the development set, we ensure that the randomly selected 3 examples from each difficulty level provided in the question enrichment prompt are related to a database different from the one associated with the current query. This approach prevents enriched question examples directly related to the database of the considered query from being included in the prompt. In the few-shot examples, we include both the enriched question and the enrichment reasoning, manually prepared considering the chain-of-thought technique, to fully leverage reasoning capabilities of the model and explicitly outline the SQL logical steps. The question enrichment prompt also includes the database schema, relevant database descriptions and values, and candidate predicates generated by the Candidate Predicate Generation (CPG) module. The fully enriched question is generated in a single prompt, without iterative refinement. Iterative enrichment is left as a potential direction for future work.

\subsection{Predicate and Error-Aware SQL Refinement Module (SR)}
\label{subsec:sr}
Since the generated SQL queries may contain minor mistakes, a common approach to address these is to use a correction module, as shown in  \autoref{fig:general-text2sql-pipeline}. Pourreza et al. ~\cite{pourreza-2023-dinsql} proposed a novel self-correction module that relies heavily on LLMs to identify and correct potential errors without executing the SQL queries. Another approach employed by C3~\cite{dong-2023-c3}, DAIL-SQL~\cite{Gao-2024-dail-sql}, CHESS~\cite{talaei2024chess}, and MCS-SQL~\cite{lee-2024-mcs-sql} is to generate multiple SQL queries for a natural language question and select the most consistent one as the final predicted SQL, rather than relying solely on the greedily generated SQL. The refiner agent proposed by Wang et al. ~\cite{wang-2024-macsql} verifies the syntactic correctness and executability of the generated SQL, triggering a correction operation if the SQL fails these checks. The refiner agent performs these actions iteratively until the result is correct or the maximum number of iterations is reached. Although this design enhances performance, it also increases both the cost and time required to produce the final SQL query. In our work, as illustrated in \autoref{fig:our-pipeline}, we execute the candidate SQL query and detect any execution errors. Using this error information, along with candidate predicates, the enriched question, and database schema, we instruct the LLM to either generate a new SQL query or refine the existing candidate SQL query.

The prompt skeletons used for each module in our pipeline are provided in our Github repository. Furthermore, a detailed example illustrating the complete workflow of the pipeline for a sample user query is also provided for reference.

\section{Experiments and Results}
\subsection{Experiment Settings}
\subsubsection{Dataset}
We conduct our experiments on the Spider~\cite{yu-etal-2018-spider} dataset and the most challenging cross-domain BIRD dataset ~\cite{li-2024-bird-sql}. Spider benchmark includes 10,181 Text-to-SQL pairs from 200 databases spanning 138 domains, each containing multiple tables. For our experiments, we used the test split of the Spider dataset, which consists of 2,147 Text-to-SQL pairs. BIRD focuses on the real-word complex and large databases, and it provides external knowledge to enhance the capability of large language models to understand both the question and the database structure and values better. BIRD dataset spans 37 professional domains such as football, formula 1, blockchain, healthcare, and education, etc., and it contains 12,751 text-to-SQL pairs from 95 databases with a size of 33.4 GB. The training set consists of 9,428 Text-to-SQL pairs, while the development and test sets consist of 1,534 and 1,789 instances, respectively. The test set of the BIRD dataset is concealed.

\subsubsection{Evaluation Metrics}
In our experiments, we evaluate model performance using the following metrics: Execution Accuracy (EX), the Reward-based Valid Efficiency Score (R-VES) and Soft F1-score, as defined by the BIRD benchmark ~\cite{li-2024-bird-sql}. Execution Accuracy (EX) measures the correctness of the predicted SQL queries by comparing their execution results to those of the ground-truth SQLs. This metric accounts for the fact that SQL queries can take multiple forms but still produce identical outcomes, thus emphasizing result accuracy over syntactic similarity. 

In addition to EX, in the latest version, the BIRD benchmark introduces a Reward-based Valid Efficiency Score (R-VES) to capture the execution efficiency of correctly predicted SQL queries. R-VES is an adjusted version of the previously proposed Valid Efficiency Score (VES). R-VES evaluates the model considering both the accuracy and the runtime performance of the SQL queries. Calculation of R-VES is provided below ~\cite{li-2024-bird-sql}.

\[
\text{R-VES} =
\begin{cases} 
1.25 & \text{if } \hat{y} \text{ is correct and } \tau \geq 2 \\
1 & \text{if } \hat{y} \text{ is correct and } 1 \leq \tau < 2 \\
0.75 & \text{if } \hat{y} \text{ is correct and } 0.5 \leq \tau < 1 \\
0.5 & \text{if } \hat{y} \text{ is correct and } 0.25 \leq \tau < 0.5 \\
0.25 & \text{if } \hat{y} \text{ is correct and } \tau < 0.25 \\
0 & \text{if } \hat{y} \text{ is incorrect}
\end{cases}
\]

Where:
\begin{itemize}
    \item $\hat{y}$ represents the predicted SQL.
    \item $\tau = \frac{\text{Ground truth SQL run time}}{\text{Predicted SQL run time}}$ represents the time ratio. $\tau$ is calculated by running the SQL 100 times, taking the average, and dropping any outliers.
\end{itemize}

Moreover, the BIRD benchmark introduced the Soft F1-score metric, designed to address the limitations of evaluation metrics such as EX. The Soft F1-score allows for a more lenient assessment by mitigating the impact of minor discrepancies, such as column reordering or missing values, in the table outputs. This makes it a complementary metric to EX, providing a more flexible evaluation of the model's performance in real-world scenarios.

\subsubsection{Models}
In this work, we employed the latest proprietary models, (GPT-4o-mini and GPT-4o) and small open-source LLMs (Qwen2.5 Coder~\cite{hui2024qwen2}, DeepSeek Coder~\cite{deepseek-coder}) without fine-tuning as the backbone for our experiments. Exploring other models or fine-tuning open-source LLMs are left as future work. For proprietary models, the majority of the experiments were conducted using GPT-4o-mini, which is approximately 30\% more cost-effective than GPT-4o. Despite its lower cost, GPT-4o-mini demonstrated robust capabilities, making it an excellent choice for balancing performance and budget constraints.

\subsubsection{Hyperparameters}
We standardized the experimental settings across all tests. The temperature was set to 0.0 to promote deterministic outputs, while the top\_p parameter, which refers to the nucleus sampling technique, was set to 1.0. The number of chat completion choices was fixed at 1; however, this could be increased in future work by employing techniques like multiple-choice selection or self-consistency. The maximum token limit was set to 2048. For each module of the E-SQL pipeline, 9-shot examples were provided, with 3 randomly selected examples for each difficulty level. Regarding database column values, the 10 most relevant distinct values to the question were selected and used. Additionally, the 20 most relevant database description sentences were identified and incorporated.

\subsection{Results}

We evaluate the proposed E-SQL pipeline primarily on the Spider test set and the BIRD development set, as the BIRD test set is not publicly available. We conduct most of our experiments using GPT-4o-mini due to its cost-effectiveness. We also experiment with small open-source models without fine-tuning to evaluate the impact of the pipeline and highlight the importance of database-aligned questions. \autoref{tab:e-sql-performance-comparison} compares the performance of E-SQL and several baseline models on the BIRD dataset, demonstrating competitive results. Additionally, we evaluate the performance of our method across various difficulty levels on BIRD dataset. \autoref{tab:e-sql-performance-comparison-spider}  provides insights into the performance of E-SQL utilizing cost-effective proprietary LLMs and small open-source LLMs on the Spider test dataset.

E-SQL shows a notable improvement, particularly for challenging questions in BIRD dataset, largely due to the incorporation of the Question Enrichment (QE) and SQL Refinement (SR) modules. With GPT-4o, the pipeline achieved an Execution Accuracy (EX) of 66.29 and a Soft F1-Score of 67.93 on the BIRD test set. When using the more cost-effective model, GPT-4o-mini, E-SQL achieved 59.81 EX and 61.59 Soft F1-Score on the BIRD test set. Additionally, using the small open-source LLM Qwen2.5 Coder 7B, E-SQL achieved 53.59 EX on the BIRD development set. On the Spider test set, E-SQL achieved 74.75 EX score using GPT-4o-mini and 58.64 EX score using Qwen2.5 Coder 7B Instruct. The detailed breakdown of performance across different datasets and query complexity levels can be found in \autoref{tab:e-sql-detailed-test-results} and \autoref{tab:e-sql-small-llm-qe-and-cpg-performance}. Further insights into the  impact of question enrichment and candidate predicate augmentation are provided in Sections \ref{sec:ablation-study} and \ref{sec:impact-of-qe-on-small-llms}. These results highlight the efficacy of question enrichment, candidate predicate augmentation, and schema refinement techniques, especially in handling complex queries.

\begin{table}[htbp]
\centering
\caption{Performance of E-SQL on BIRD development and test set. E-SQL consists of CSG, CPG, QE, and SR modules as illustrated in \autoref{fig:our-pipeline}.}
\label{tab:e-sql-performance-comparison}
\resizebox{\columnwidth}{!}{% Scale to fit the column
\begin{tabular}{lccc}
\toprule
\textbf{Method} & \textbf{Dev EX} & \textbf{Test EX}  & \textbf{Test R-VES}\\
\midrule
\multicolumn{4}{c}{\textbf{Undisclosed}}\\
\midrule
OpenSearch-SQL, v2 + GPT-4o  & 69.30 & \textbf{72.28} & \textbf{69.36}\\
Distillery + GPT-4o & 67.21 & 71.83 & 67.41\\
ExSL + granite-34b-code & 67.47 & 70.37 & 68.79\\
Insights AI & \textbf{72.16} & 70.26 & 66.39 \\
PURPLE + RED + GPT-4o & 68.12 & 70.21 & 65.62\\
ByteBrain & 65.45 & 68.87 & -\\
ExSL + granite-20b-code & 65.38	 & 67.86 & 66.25\\
Arcwise + GPT-4o & 67.99 & 66.21 & 63.68\\
SCL-SQL & 64.73 & 65.23 & 61.28 \\
OpenSearch-SQL,v1 + GPT-4 & 61.34 & 64.95 & - \\
PB-SQL, v1 & 60.50 & 64.84 & 60.36\\
\midrule
\multicolumn{4}{c}{\textbf{Published}}\\
\midrule
CHESS & 65.00 & 66.69 & 62.77\\
% RSL-SQL + DeepSeek & 63.56 & 65.51 & -\\ %not in the leaderbord at submission
MCS-SQL + GPT-4 ~\cite{lee-2024-mcs-sql} & 63.36 & 65.45 & 61.23\\
SuperSQL ~\cite{li2024SuperSQL} & 58.50 & 62.66 & - \\
SFT CodeS-15B ~\cite{li-2024-codes} & 58.47 & 60.37 & 61.37 \\
DTS-SQL + DeepSeek 7B ~\cite{Pourreza-2024-dts-sql} & 55.80 & 60.31 & - \\
MAC-SQL + GPT-4 ~\cite{wang-2024-macsql} & 57.56 & 59.59 & 57.60 \\
SFT CodeS-7B ~\cite{li-2024-codes} & 57.17 & 59.25 & 55.69 \\
TA-SQL + GPT-4 ~\cite{qu-etal-2024-ta-sql} & 56.19 & 59.14 & - \\
DAIL-SQL + GPT-4 ~\cite{Gao-2024-dail-sql} & 54.76 & 57.41 & 54.02 \\
\midrule
\textbf{E-SQL + GPT-4o (Ours)} & 65.58 & 66.29 & 62.43 \\
\textbf{E-SQL + GPT-4o-mini (Ours)} & 61.60 & 59.81 & 55.64\\
\textbf{E-SQL + Qwen2.5 Coder 7B Instruct (Ours)} & 53.59 & - & -\\
\bottomrule
\end{tabular}
}
\end{table}

\begin{table}[htbp]
\centering
\caption{Performance of E-SQL on Spider test set. E-SQL consists of CSG, CPG, QE, and SR modules as illustrated in \autoref{fig:our-pipeline}. \textsuperscript{\dag} symbol denotes methods utilizing fine-tuned LLMs.}
\label{tab:e-sql-performance-comparison-spider}
\resizebox{0.90\linewidth}{!}{% Scale to fit the column
\scriptsize % Use smaller font
\begin{tabular}{lcc}
\toprule
\textbf{Method} & \textbf{Model} & \textbf{Test EX} \\
\midrule
\multicolumn{3}{c}{\textbf{Proprietary Models}}\\
\midrule

DAIL-SQL~\cite{Gao-2024-dail-sql} & GPT-4 & 86.6 \\
DIN-SQL~\cite{pourreza-2023-dinsql} & GPT-4 & 85.3 \\
TA-SQL~\cite{qu-etal-2024-ta-sql} & GPT-4 & 85.0 \\
MAC-SQL~\cite{wang-2024-macsql} & GPT-3.5-Turbo & 75.5 \\
- & GPT4 & 74.0 \\
\midrule
\textbf{E-SQL (ours)} &GPT-4o-mini & 74.75 \\

\midrule
\multicolumn{3}{c}{\textbf{Small Open Source Models}}\\
\midrule
DTS-SQL\textsuperscript{\dag}~\cite{Pourreza-2024-dts-sql} & Mistral-7B & 77.1 \\
MSc-SQL\textsuperscript{\dag}~\cite{msc-sql-multisamplecritiquingsmall} & Gemma-2-9B & 69.30\\

\midrule
\textbf{E-SQL (ours)} &Qwen2.5 Coder 7B Instruct  & 58.64 \\
\bottomrule
\end{tabular}
}
\end{table}

\begin{table*}[htbp]
\centering
\caption{Detailed Performance of E-SQL on BIRD Test Set Across Query Complexity Levels}
\label{tab:e-sql-detailed-test-results}
\resizebox{\textwidth}{!}{ % This scales the table to fit the width of the page
\begin{tabular}{lcccccccccccc}
\toprule
\multirow{2}{*}{\textbf{Pipeline}} & \multicolumn{3}{c}{\textbf{Overall}} & \multicolumn{3}{c}{\textbf{Simple}} & \multicolumn{3}{c}{\textbf{Moderate}} & \multicolumn{3}{c}{\textbf{Challenging}} \\
\cmidrule(lr){2-4} \cmidrule(lr){5-7} \cmidrule(lr){8-10} \cmidrule(lr){11-13}
 & \textbf{EX} & \textbf{Soft F1} & \textbf{R-VES} & \textbf{EX} & \textbf{Soft F1} & \textbf{R-VES} & \textbf{EX} & \textbf{Soft F1} & \textbf{R-VES} & \textbf{EX} & \textbf{Soft F1} & \textbf{R-VES} \\
\midrule
\textbf{E-SQL (GPT-4o)}  & 66.29 & 67.93 & 62.43 & 73.02 & 73.91 &68.68 & 64.14 & 66.17 &60.46 & 48.07 & 51.45 & 54.48 \\
\textbf{E-SQL (GPT-4o-mini)}  & 59.81 & 61.59 & 55.64 & 67.44 & 68.80 & 62.53 & 56.94 & 58.77 & 53.11 & 40.00 & 43.04 & 37.60 \\
% \textbf{E-SQL (Qwen2.5 Coder 7B Instruct)}  & 53.59 & 56.11 & - & 60.76 & 62.86 & - & 43.10 & 45.99 & - & 41.38 & 45.40 & - \\
\bottomrule
\end{tabular}
}
\end{table*}

\begin{table*}[htbp]
\centering
\caption{Performance of different pipelines on BIRD development set. SF represents the Schema Filtering module, QE represents the Basic Question Enrichment module, and G represents the Basic SQL Generation module. The arrows indicate performance improvement (\(\uparrow\)) or decline (\(\downarrow\)) compared to the G baseline. Experiments are conducted using GPT-4o-mini.}
\label{tab:SF-performance}
\begin{tabular}{lcccc}
\toprule
\textbf{Pipeline} & \textbf{Overall Dev EX} & \textbf{Simple EX} & \textbf{Moderate EX} & \textbf{Challenging EX} \\
\midrule
SF-G     & 49.48 $(\downarrow 8.21)$  & 58.16 $(\downarrow 6.38)$  & 36.85 $(\downarrow 12.28)$ & 34.48 $(\downarrow 8.96)$ \\
SF-QE-G  & 55.34 $(\downarrow 2.35)$  & 62.27 $(\downarrow 2.27)$  & 46.12 $(\downarrow 3.01)$  & 40.68 $(\downarrow 2.76)$ \\
QE-G  & 58.80 $(\uparrow 1.11)$ & 64.43 $(\downarrow 0.11)$ & 51.07 $(\uparrow 1.94)$ & 47.58 $(\uparrow 4.14)$ \\
\midrule
G & 57.69 & 64.54 & 49.13 & 43.44 \\
\bottomrule
\end{tabular}
\end{table*}

\begin{table*}[htbp]
\centering
\caption{Ablation study using GPT-4o-mini with EX and Soft F1 metrics on the BIRD development set. The arrows indicate performance improvement (\(\uparrow\)) or decline (\(\downarrow\)) compared to the base E-SQL.}
\label{tab:ablation-study-gpt-4o-mini}
\resizebox{\textwidth}{!}{ % This scales the table to fit the width of the page
\begin{tabular}{lcccccccc}
\toprule
\multirow{2}{*}{\textbf{Pipeline}} & \multicolumn{2}{c}{\textbf{Overall}} & \multicolumn{2}{c}{\textbf{Simple}} & \multicolumn{2}{c}{\textbf{Moderate}} & \multicolumn{2}{c}{\textbf{Challenging}} \\
\cmidrule(lr){2-3} \cmidrule(lr){4-5} \cmidrule(lr){6-7} \cmidrule(lr){8-9}
 & \textbf{EX} & \textbf{Soft F1} & \textbf{EX} & \textbf{Soft F1} & \textbf{EX} & \textbf{Soft F1} & \textbf{EX} & \textbf{Soft F1} \\
\midrule
\textbf{E-SQL}  & 61.60 & 65.61 & 68.00 & 71.54  & 53.23 & 58.34 & 47.59 & 51.02 \\
\midrule
\textbf{w/o QE}  & 59.71 $(\downarrow 1.89)$ & 63.84 $(\downarrow 1.77)$ & 66.05 $(\downarrow 1.95)$ & 69.86 $(\downarrow 1.68)$ & 52.37 $(\downarrow 0.86)$ & 57.27 $(\downarrow 1.07)$  & 42.75 $(\downarrow 4.84)$ & 46.52 $(\downarrow 4.50)$ \\
\textbf{w/o CPG}  & 59.58 $(\downarrow 2.02)$ & 63.61 $(\downarrow 2.00)$ & 65.51 $(\downarrow 2.49)$ & 69.16 $(\downarrow 2.38)$ & 51.29 $(\downarrow 1.94)$ & 56.27 $(\downarrow 2.07)$ & 48.27 $(\uparrow 0.68)$ & 51.68 $(\uparrow 0.66)$ \\
\textbf{w/o QE \& CPG}  & 58.34 $(\downarrow 3.26)$ & 62.41 $(\downarrow 3.20)$ & 64.22 $(\downarrow 3.78)$ & 67.91 $(\downarrow 3.63)$ & 51.29 $(\downarrow 1.94)$ & 55.66 $(\downarrow 2.68)$ & 43.45 $(\downarrow 4.14)$ & 48.89 $(\downarrow 2.13)$ \\
\textbf{w/o SR (w/o QE \& CPG \& SR)} & 58.03 $(\downarrow 3.57)$ & 61.88 $(\downarrow 3.73)$ & 63.89 $(\downarrow 4.11)$ & 67.33 $(\downarrow 4.21)$ & 50.86 $(\downarrow 2.37)$ & 55.13 $(\downarrow 3.21)$ & 44.13 $(\downarrow 3.46)$ & 48.71 $(\downarrow 2.31)$ \\
\textbf{w/ SF}  & 56.06 $(\downarrow 5.54)$ & 59.93 $(\downarrow 5.68)$ & 62.70 $(\downarrow 5.3)$ & 66.53 $(\downarrow 5.01)$ & 47.63 $(\downarrow 5.60)$ & 51.55 $(\downarrow 6.79)$  & 40.68 $(\downarrow 6.91)$ & 44.62 $(\downarrow 6.40)$ \\
\bottomrule
\end{tabular}
}
\end{table*}

\subsection{Schema Filtering}
\label{sec:sf-experiment-discussion}

Schema filtering is a technique aimed at minimizing the model's reliance on irrelevant database schema elements by removing such items, and has been applied in numerous previous works. To evaluate the impact of schema filtering on a basic pipeline utilizing most advanced large language models, we conducted various experiments both with and without a schema filtering module, which includes a filtered schema correction step. The results of these experiments are shown in \autoref{tab:SF-performance}. As demonstrated in the \autoref{tab:SF-performance}, the effect of schema filtering varies depending on its placement within the pipeline. Nevertheless, regardless of its position, incorporating the schema filtering module in a basic pipeline consistently led to a performance decline across all difficulty levels. Detailed results of the question enrichment module are discussed in the next section.

The inclusion of schema filtering resulted in an overall drop of up to 8.21\% in Execution Accuracy (EX) on the development set. Specifically, the performance decreased by 6.38\%, 12.28\%, and 8.96\% for simple, moderate, and challenging questions, respectively as shown in \autoref{tab:SF-performance}. Although the E-SQL pipeline does not inherently include a schema filtering module, we integrated it into the pipeline for our ablation study, as presented in \autoref{tab:ablation-study-gpt-4o-mini}. This integration resulted in a 5.54 drop in EX and a 5.68 decrease in Soft F1 score on the development set. These findings align with previous research ~\cite{maamari2024deathschemalinkingtexttosql-distyl-ai}, which suggests that advanced LLMs can manage schema linking effectively without requiring explicit filtering. Consequently, schema filtering was excluded from the final E-SQL pipeline, as its negative impact outweighed the potential benefits.

\subsection{Ablation Study}
\label{sec:ablation-study}
We conducted an ablation study to analyze the contributions of the Question Enrichment, Candidate Predicate Augmentation, and SQL Refinement modules. The results of this study are summarized in \autoref{tab:ablation-study-gpt-4o-mini}, showing the impact of individual modules from the E-SQL pipeline.

\begin{table}[htbp]
\centering
\caption{Effect of QE and CPG modules in E-SQL pipeline with small open-source LLMs. }
\label{tab:e-sql-small-llm-qe-and-cpg-performance}
\resizebox{\columnwidth}{!}{% Scale to fit the column
\begin{tabular}{lccc}
\toprule
\textbf{Pipeline} & \textbf{Model} & \textbf{ Spider Test EX} & \textbf{ Bird Dev EX} \\
\midrule
E-SQL & Qwen 2.5 Coder 7B & 58.64 & 53.52 \\
\midrule
w/o QE & Qwen 2.5 Coder 7B & 55.84 $(\downarrow 2.80)$ & 50.78 $(\downarrow 2.74)$\\
w/o CPG & Qwen 2.5 Coder 7B & 57.10 $(\downarrow 1.54)$ & 50.72 $(\downarrow 2.80)$ \\
w/o SR & Qwen 2.5 Coder 7B & 57.14 $(\downarrow 1.50)$ & 48.24 $(\downarrow 5.28)$ \\
\bottomrule
\end{tabular}
}
\end{table}

\subsubsection{Question Enrichment}
The Question Enrichment module, which facilitates direct schema linking by injecting database items, SQL components, conditions, and SQL generation steps into the question, improved the performance as shown in \autoref{tab:ablation-study-gpt-4o-mini} and \autoref{tab:e-sql-small-llm-qe-and-cpg-performance} on both advance proprietary and small open-source LLMs. Its impact was particularly significant on challenging questions. The absence of the question enrichment technique, especially when combined with the removal of the Candidate Predicate Augmentation module (CPG), led to a further decrease in overall performance. These results demonstrate that direct schema linking, achieved through question reformulation, effectively bridges the gap between the natural language query and the database schema, resulting in more accurate SQL generation.

\subsubsection{Possible Predicate Augmentation}
The Candidate Predicate Augmentation (CPG) module enhances the pipeline by augmenting potential predicates extracted from the database with the help of the \texttt{LIKE} operator and the candidate SQL query. As shown in the \autoref{tab:ablation-study-gpt-4o-mini} and \autoref{tab:e-sql-small-llm-qe-and-cpg-performance}, removing the CPG module resulted in a nearly 2\% drop in overall model performance. However, its removal slightly improved the performance on challenging questions of BIRD dev set, suggesting that the CPG module may introduce unnecessary complexity in some cases. The slight negative effect of the CPG module on challenging questions is negligible, as it substantially enhances overall performance, especially when compared to the significant gains achieved through the Question Enrichment module.

\subsubsection{SQL Refinement}
The SQL Refinement (SR) module plays a crucial role in correcting minor errors in the generated SQL queries. Without SR, we observed a 3.57 drop in EX and a 3.73 decrease in Soft F1 across the BIRD development set. This demonstrates that the SQL refinement step significantly boosts the final query accuracy by detecting and correcting SQL execution errors.

To further evaluate the impact of the SR module within the E-SQL pipeline, we conducted the following analyses, with results presented in \autoref{tab:sr-analysis}:
\begin{itemize}
    \item The proportion of initially generated candidate SQL queries that were altered by the SR module. A candidate SQL query is counted as changed if the final predicted SQL query differs from the original candidate SQL. 
    \item  The proportion of initially non-executable candidate SQL queries that were modified by the SR module to become executable.
    \item The proportion of non-executable candidate SQL queries that were corrected to executable and accurate SQL queries by the SR module.
    \item The proportion of incorrect candidate SQL queries, including non-executable ones, that were corrected to accurate SQL queries by the SR module.
\end{itemize}

As shown in \autoref{tab:sr-analysis}, our detailed analysis of the SQL Refinement (SR) module demonstrates that 5.35\% and 1.83\% of the initially incorrect SQL queries generated by GPT-4o-mini and GPT-4o, respectively, were successfully corrected by the SR module. While the SQL refinement technique positively impacts both models, the effect is more noticeable on less capable models like GPT-4o-mini. These results highlight the module's ability to enhance query accuracy, especially for models with lower initial performance, making SQL refinement a critical component for improving the overall system robustness.

\begin{table}[htbp]
\centering
\caption{Analysis of SQL Refinement (SR) Module on the BIRD Development Set}
\label{tab:sr-analysis}
\resizebox{0.90\linewidth}{!}{
  \begin{tabular}{lcc}
    \toprule
    \multirow{2}{*}{\textbf{Metric}} & \multicolumn{2}{c}{\textbf{E-SQL SR}} \\
    \cmidrule(lr){2-3}
    & \textbf{GPT-4o-mini} & \textbf{GPT-4o} \\
    \midrule
    Changed Queries (\%) & 49.48 & 23.20 \\
    Non-Executable to Executable (\%) & 6.58 & 0.39 \\
    Non-Executable to Correct (\%) & 3.19 & 0.13 \\
    Wrong to Correct (\%) & 5.35 & 1.83 \\
    \bottomrule
  \end{tabular}
}
\end{table}

\subsection{Impact of Enriched Questions on Small Large Language Models}
\label{sec:impact-of-qe-on-small-llms}
To evaluate the performance impact of database-integrated questions on small LLMs ~\cite{qwen2, hui2024qwen2, deepseek-coder}, we conducted an experiment comparing their performance on default questions versus enriched questions.  In this experiment, SQL queries for a given question were generated using a single-prompt approach. The prompt template used in this step is similar to that of the Candidate SQL Generation (CSG) module in the E-SQL pipeline, excluding data augmentation components such as database descriptions, value samples, and few-shot examples. This approach allows us to isolate the effect of enriched questions, as the single prompt relies solely on instructions and questions without additional context.  Enriched questions were extracted from the outputs of Question Enrichment Module (QE) of the E-SQL pipeline utilized with GPT-4o to ensure that they incorporated database items and SQL construction plans. 

The results, presented in ~\autoref{tab:enriched-question-effect-on-small-os-llms}, indicate that even small LLMs can achieve competitive performance without task-specific fine-tuning when provided with high-quality, database-integrated natural language queries. These findings underscore the critical role of database-integrated enriched questions, which include logical SQL construction steps.

\begin{table}[htbp]
\centering
\caption{Effect of Enriched Questions on the performance of small open-source LLMs without fine-tuning on BIRD development set. Enriched questions, generated using GPT-4o, were utilized to evaluate the impact of high-quality question enrichment on the performance of small open-source models. \textsuperscript{\dag} symbol denotes methods utilizing fine-tuned LLMs.}
\label{tab:enriched-question-effect-on-small-os-llms}
\resizebox{\columnwidth}{!}{ % This scales the table to fit the width of the page
\begin{tabular}{lcccccccc}
\toprule
\multirow{2}{*}{\textbf{Model}} & \multirow{2}{*}{\textbf{Level}} & \multicolumn{2}{c}{\textbf{Dev EX}} \\
\cmidrule{3-4}
& & \textbf{Default} & \textbf{Enriched} \\
\midrule
\multirow{4}{*}{\textbf{DeepSeek Coder 1.3B Instruct}} &Overall & 20.92 & 50.84$(\uparrow 29.92)$  \\
 &Simple  & 28.43 & 62.38$(\uparrow 33.95)$   \\
 &Moderate  & 10.34 & 36.42$(\uparrow 26.08)$    \\
 &Challenging  & 6.90 & 23.45$(\uparrow 16.55)$  \\
\midrule
\multirow{4}{*}{\textbf{Qwen2.5 Coder 1.5B Instruct}} & Overall & 11.21 & 36.90 $(\uparrow 25.69)$  \\
 &Simple  & 14.27 & 44.10$(\uparrow 29.83)$   \\
 &Moderate  & 6.68 & 28.23$(\uparrow 21.55)$    \\
 &Challenging  & 6.20 & 18.62$(\uparrow 12.42)$  \\
\midrule
\multirow{4}{*}{\textbf{DeepSeek Coder 7B Instruct 1.5v}} &Overall & 37.02 & 56.45$(\uparrow 19.43)$  \\
 &Simple  & 44.65 &  64.64$(\uparrow 19.99)$  \\
 &Moderate  & 26.52  & 45.47$(\uparrow 18.95)$    \\
 &Challenging  & 22.06 & 39.31$(\uparrow 17.25)$  \\
\midrule
\multirow{4}{*}{\textbf{Qwen2.5 Coder 7B Instruct}} &Overall & 31.25 & 40.22$(\uparrow 8.97)$  \\
 &Simple  & 40.43 & 50.70$(\uparrow 10.27)$   \\
 &Moderate  & 17.88 & 26.07$(\uparrow 8.19)$    \\
 &Challenging  & 15.17 & 18.62$(\uparrow 3.45)$  \\
\midrule
\textbf{ExSL + granite-20b-code
} & Overall & 51.69 & - \\
\textbf{DTS-SQL + DeepSeek 7B \textsuperscript{\dag}}\cite{Pourreza-2024-dts-sql} & Overall & 55.8 & - \\
\textbf{SFT CodeS-7B \textsuperscript{\dag}} ~\cite{li-2024-codes} & Overall & 57.17 & - \\
\textbf{SFT CodeS-15B \textsuperscript{\dag}} ~\cite{li-2024-codes}  & Overall & 58.47 & - \\
\bottomrule
\end{tabular}
}
\end{table}

\subsection{Computational Cost Analysis}

% Prompt templates contain the following number of tokens by default. Note that the token numbers are found using GPT-4o tokenizer and GPT-4o-mini tokenizer
% -- Candidate SQL Generation Prompt Template: 593 tokens
% -- Question Enrichment Prompt Template: 801 tokens
% -- Schema Filtering Prompt Template: 725 tokens
% -- SQL Refinement Prompt Template: 901 tokens

Understanding computational expenses is critical for assessing the practical scalability and applicability of the framework, especially given the reliance on large language models (LLMs) and their resource demands. Table~\ref{tab:question_token_no} highlights the impact of question enrichment on token count\textsuperscript{2}, while Table~\ref{tab:computational-cost} provides details of average token usage for both prompt and completion stages of key pipeline components on the BIRD development set. The average number of tokens in prompts is inherently high due to well-defined instructions, data augmentations, including few-shot examples, database descriptions, and database values, as commonly employed in most Text-to-SQL approaches. Consequently, the increase in the question token count due to enrichment is relatively insignificant compared to the total number of prompt tokens. Despite the computational overhead introduced by question enrichment, which increases the average token count of natural language questions and their reasoning, E-SQL demonstrates computational superiority over methods ~\cite{dong-2023-c3, lee-2024-mcs-sql, talaei2024chess, Gao-2024-dail-sql, maamari2024deathschemalinkingtexttosql-distyl-ai} that generate multiple SQL queries for a single user question. These methods incur significant computational costs due to the repeated SQL generation, subsequent correction of these SQL queries, and the selection process. Executing each E-SQL module only once minimizes the computational costs associated with repeated calls, ensuring greater efficiency. This balance between cost and performance underscores the scalability and efficiency of our pipeline for large-scale deployment. 

\footnotetext[2]{Token counts were computed using the tiktoken Python package, developed by OpenAI, which provides a programmatic interface for tokenizing text with OpenAI model-specific tokenizers. The package is available at [https://github.com/openai/tiktoken].}

\begin{table}[htbp]
\centering
\caption{Comparison of Default and Enriched Natural Language Questions in Bird Development Set}
\label{tab:question_token_no}
\resizebox{0.80\linewidth}{!}{
  \begin{tabular}{lc}
    \toprule
    \textbf{Text} & \textbf{Avg. Tokens}  \\
    \midrule
    \textbf{Question} & 18.36 \\
    \textbf{Enriched Question} & 81.51 \\
    \textbf{Enrichment Reasoning} & 191.34 \\
    \textbf{Fully Enriched Question} & 291.21 \\
    \bottomrule
  \end{tabular}
}
\end{table}

\begin{table}[htbp]
\centering
\caption{Analysis of Computational Costs by Module on the Bird Development Set}
\label{tab:computational-cost}
\resizebox{0.80\linewidth}{!}{
  \begin{tabular}{lccc}
    \hline
    \multirow{2}{*}{\textbf{Module}} & 
    \textbf{Avg. Prompt} & 
    \textbf{Avg. Completion} \\
    & \textbf{Token Count} & \textbf{Token Count} \\
    \hline
    \textbf{CSG} & 12612 & 199 \\
    \textbf{QE}  & 16550 & 292 \\
    \textbf{SR}  & 7403  & 267 \\
    \hline
  \end{tabular}
}
\end{table}

In our pipeline, the initially generated candidate SQL query is executed to identify execution errors and enhance the large language model's error awareness. While this step contributes to the overall pipeline latency, with an average execution time of 49.936 milliseconds per query, it plays a crucial role in ensuring accurate SQL refinement by providing valuable feedback on execution errors. It is important to note that the overall pipeline response time varies based on several factors, including the complexity of the natural language question, the length of the prompt, and the API response time of proprietary LLMs, which is influenced by server load and volume. These factors collectively contribute to the latency of the system.

\section{Discussion and Limitations}
The results from our experiments highlight the significant influence of question enrichment and candidate predicate augmentation on the performance of the E-SQL pipeline. The question enrichment module, which bridges the gap between the natural language query and the database schema, was pivotal in improving query accuracy, particularly for challenging questions. By enriching the natural language question with database items, conditions, and SQL generation steps, the module enhanced direct schema linking, ensuring that the generated SQL queries were more aligned with the database's structure. This improvement is evidenced by an ablation study, underscoring the efficacy of this approach.

One notable observation in our evaluation is the inconsistency in performance between the development and test sets when using different models. Specifically, when employing GPT-4o, the pipeline's performance showed an improvement on the test set compared to the development set. However, this trend reversed with GPT-4o-mini, where performance decreased on the test set relative to the development set. Due to the BIRD test set not being publicly available, we were unable to analyze it directly to identify potential causes for this variation. Additionally, large language models are known to exhibit variability in performance across multiple runs, which might further contribute to this inconsistency. Thus, while the exact reasons behind these performance fluctuations remain unclear, they underline the need for further exploration under controlled conditions.

The prompt design plays a critical role in influencing model performance. The prompt templates utilized for each module of the E-SQL pipeline are publicly available in our GitHub repository. This study primarily emphasizes schema linking through question enrichment and data augmentation, deliberately leaving the exploration of alternative prompt templates beyond its scope.

Despite the advancements, there are some limitations to our approach. Due to hardware and cost constraints, almost all experiments were conducted using GPT-4o-mini and small open-source LLMs without fine-tuning. Among the small open-source LLMs, the whole E-SQL pipeline was executed only with Qwen2.5 Coder 7B Instruct since the context length of the other small LLMs is not sufficient to run and observe the effect of E-SQL pipeline. Developing more efficient schema linking techniques that operate effectively with small LLMs and limited context lengths represents a promising direction for future work.

\section{Conclusions}
In this study, we introduced E-SQL, a novel pipeline designed to address key challenges in Text-to-SQL translation by leveraging direct schema linking via question enrichment and incorporating candidate predicates. Our experiments demonstrated that the question enrichment module, which integrates natural language queries with relevant database elements and logical steps, significantly enhances query accuracy, particularly for complex queries. Additionally, the proposed candidate predicate augmentation technique further improves the performance of the pipeline. Moreover, our additional experiments reveals the importance and positive impact of enriched questions on the performance of small open-source LLMs with limited context lengths.

While some prior works have highlighted the utility of schema filtering, our findings reveal that incorporating schema filtering into a text-to-SQL translation pipeline that leverages advanced LLMs results in performance degradation. This supports the notion that explicit schema filtering can be redundant in modern architectures that utilizes the latest LLMs. 

By focusing on question enrichment, data augmentation and SQL refinement, E-SQL achieved competitive results on the BIRD benchmark. Specifically, E-SQL combined with GPT-4o achieved 65.58\% and 66.29\% execution accuracy on the development and test sets, respectively. These results underscore E-SQL's effectiveness in handling complex queries and present it as a promising approach for future Text-to-SQL tasks.

Despite a minor computational overhead due to increased token counts in question enrichment, its impact is negligible compared to the overall token usage and is outweighed by the significant performance gains. Additionally, E-SQL ensures cost-efficiency by executing each module only once, avoiding the excessive resource demands of repeated query generation and correction. This balance highlights E-SQL's scalability and suitability for resource-constrained deployments.

Further exploration of fine tuning, iterative or multiple question refinements and schema linking techniques optimized for small LLMs with limited context lengths is left for future work.

\begin{acks}
 We would like to express our sincere gratitude to Dr. Arif Usta and Ekrem Polat for their invaluable insights and constructive discussions, which greatly contributed to the development of this work.
\end{acks}

\bibliographystyle{ACM-Reference-Format}
\bibliography{sample-base}

\appendix
\onecolumn % Switch to single-column mode for the appendix

\section{Prompt Templates}
\label{appendix:A}
In this section, exact prompt templates used for each module in the E-SQL pipeline are provided. 

\subsection{Full Prompt Template for Candidate SQL Generation (CSG)}
\label{appendix:A1}

\begin{tcolorbox}[mybluebox, width=\textwidth]
\lstset{
    breaklines=true,
    basicstyle=\ttfamily\footnotesize, % Adjusting to a smaller font size
    columns=fullflexible,
    frame=none,
    backgroundcolor=\color{blue!5!white},
    xleftmargin=0pt,
    xrightmargin=0pt,
    showspaces=false,
    showstringspaces=false,
    keepspaces=true, % Keep spaces exactly as they are, without any additional indentation
    breakindent=0pt, % No indent for wrapped lines
    aboveskip=0pt,
    belowskip=0pt,
}

\begin{lstlisting}
### You are an excellent data scientist. You can capture the link between the question and corresponding database and perfectly generate valid SQLite SQL query to answer the question. Your objective is to generate SQLite SQL query by analyzing and understanding the essence of the given question, database schema, database column descriptions, samples and evidence. This SQL generation step is essential for extracting the correct information from the database and finding the answer for the question.

### Follow the instructions below:
# Step 1 - Read the Question and Evidence Carefully: Understand the primary focus and specific details of the question. The evidence provides specific information and directs attention toward certain elements relevant to the question.
# Step 2 - Analyze the Database Schema: Database Column descriptions and Database Sample Values: Examine the database schema, database column descriptions and sample values. Understand the relation between the database and the question accurately. 
# Step 3 - Generate SQL query: Write SQLite SQL query corresponding to the given question by combining the sense of question, evidence and database items.

{FEWSHOT_EXAMPLES}

### Task: Given the following question, database schema and evidence, generate SQLite SQL query in order to answer the question.
### Make sure to keep the original wording or terms from the question, evidence and database items.
### Make sure each table name and column name in the generated SQL is enclosed with backtick separately.
### Ensure the generated SQL is compatible with the database schema.
### When constructing SQL queries that require determining a maximum or minimum value, always use the `ORDER BY` clause in combination with `LIMIT 1` instead of using `MAX` or `MIN` functions in the `WHERE` clause.Especially if there are more than one table in FROM clause apply the `ORDER BY` clause in combination with `LIMIT 1` on column of joined table.
### Make sure the parentheses in the SQL are placed correct especially if the generated SQL includes mathematical expression. Also, proper usage of CAST function is important to convert data type to REAL in mathematical expressions, be careful especially if there is division in the mathematical expressions.
### Ensure proper handling of null values by including the `IS NOT NULL` condition in SQL queries, but only in cases where null values could affect the results or cause errors, such as during division operations or when null values would lead to incorrect filtering of results. Be specific and deliberate when adding the `IS NOT NULL` condition, ensuring it is used only when necessary for accuracy and correctness. This is crucial to avoid errors and ensure accurate results.  This is crucial to avoid errors and ensure accurate results. You can leverage the database sample values to check if there could be potential null value.

{SCHEMA}
{DB_DESCRIPTIONS}
{DB_SAMPLES}
{QUESTION}
{EVIDENCE}

### Please respond with a JSON object structured as follows:

{"chain_of_thought_reasoning":  "Explanation of the logical analysis and steps that result in the final SQLite SQL query.", "SQL": "Generated SQL query as a single string"}

Let's think step by step and generate SQLite SQL query.
\end{lstlisting}
\end{tcolorbox}

\clearpage
\subsection{Full Prompt Template for Quesiton Enrichment (QE)}
\label{appendix:A2}

\begin{tcolorbox}[mybluebox, width=\textwidth]
\lstset{
    breaklines=true,
    basicstyle=\ttfamily\footnotesize, % Adjusting to a smaller font size
    columns=fullflexible,
    frame=none,
    backgroundcolor=\color{blue!5!white},
    xleftmargin=0pt,
    xrightmargin=0pt,
    showspaces=false,
    showstringspaces=false,
    keepspaces=true, % Keep spaces exactly as they are, without any additional indentation
    breakindent=0pt, % No indent for wrapped lines
    aboveskip=0pt,
    belowskip=0pt,
}

\begin{lstlisting}
### You are excellent data scientist and can link the information between a question and corresponding database perfectly. Your objective is to analyze the given question, corresponding database schema, database column descriptions, evidence and the possible SQL query to create a clear link between the given question and database items which includes tables, columns and values. With the help of link, rewrite new versions of the original question to be more related with database items, understandable, clear, absent of irrelevant information and easier to translate into SQL queries. This question enrichment is essential for comprehending the question's intent and identifying the related database items. The process involves pinpointing the relevant database components and expanding the question to incorporate these items.

### Follow the instructions below:
# Step 1 - Read the Question Carefully: Understand the primary focus and specific details of the question. Identify named entities (such as organizations, locations, etc.), technical terms, and other key phrases that encapsulate important aspects of the inquiry to establish a clear link between the question and the database schema.
# Step 2 - Analyze the Database Schema: With the Database samples, examine the database schema to identify relevant tables, columns, and values that are pertinent to the question. Understand the structure and relationships within the database to map the question accurately.
# Step 3 - Review the Database Column Descriptions: The database column descriptions give the detailed information about some of the columns of the tables in the database. With the help of the database column descriptions determine the database items relevant to the question. Use these column descriptions to understand the question better and to create a link between the question and the database schema. 
# Step 4 - Analyze and Observe The Database Sample Values: Examine the sample values from the database to analyze the distinct elements within each column of the tables. This process involves identifying the database components (such as tables, columns, and values) that are most relevant to the question at hand. Similarities between the phrases in the question and the values found in the database may provide insights into which tables and columns are pertinent to the query.
# Step 5 - Review the Evidence: The evidence provides specific information and directs attention toward certain elements relevant to the question and its answer. Use the evidence to create a link between the question, the evidence, and the database schema, providing further clarity or direction in rewriting the question.
# Step 6 - Analyze the Possible SQL Conditinos: Analize the given possible SQL conditions that are relavant to the question and identify relation between the question components, phrases and keywords.
# Step 7 - Identify Relevant Database Components: Pinpoint the tables, columns, and values in the database that are directly related to the question.
# Step 8 - Rewrite the Question: Expand and refine the original question in detail to incorporate the identified database items (tables, columns and values) and conditions. Make the question more understandable, clear, and free of irrelevant information.

{FEWSHOT_EXAMPLES}

### Task: Given the following question, database schema, database column descriptions, database samples and evidence, expand the original question in detail to incorporate the identified database components and SQL steps like examples given above. Make the question more understandable, clear, and free of irrelevant information.
### Ensure that question is expanded with original database items. Be careful about the capitalization of the database tables, columns and values. Use tables and columns in database schema.

{SCHEMA}
{DB_DESCRIPTIONS}
{DB_SAMPLES}
{POSSIBLE_CONDITIONS}
{QUESTION}
{EVIDENCE}

### Please respond with a JSON object structured as follows:

```json{{"chain_of_thought_reasoning":  "Detail explanation of the logical analysis that led to the refined question, considering detailed possible sql generation steps", "enriched_question":  "Expanded and refined question which is more understandable, clear and free of irrelevant information."}}```

Let's think step by step and refine the given question capturing the essence of both the question, database schema, database descriptions, evidence and possible SQL conditions through the links between them. If you do the task correctly, I will give you 1 million dollars. Only output a json as your response.
\end{lstlisting}
\end{tcolorbox}

\clearpage
\subsection{Full Prompt Template for SQL Refinement (SR)}
\label{appendix:A3}

\begin{tcolorbox}[mybluebox, width=\textwidth]
\lstset{
    breaklines=true,
    basicstyle=\ttfamily\footnotesize, % Adjusting to a smaller font size
    columns=fullflexible,
    frame=none,
    backgroundcolor=\color{blue!5!white},
    xleftmargin=0pt,
    xrightmargin=0pt,
    showspaces=false,
    showstringspaces=false,
    keepspaces=true, % Keep spaces exactly as they are, without any additional indentation
    breakindent=0pt, % No indent for wrapped lines
    aboveskip=0pt,
    belowskip=0pt,
}

\begin{lstlisting}
### You are an excellent data scientist. You can capture the link between the question and corresponding database and perfectly generate valid SQLite SQL query to answer the question. Your objective is to generate SQLite SQL query by analyzing and understanding the essence of the given question, database schema, database column descriptions, evidence, possible SQL and possible conditions. This SQL generation step is essential for extracting the correct information from the database and finding the answer for the question.

### Follow the instructions below:
# Step 1 - Read the Question and Evidence: Understand the primary focus and specific details of the question. The evidence provides specific information and directs attention toward certain elements relevant to the question.
# Step 2 - Analyze the Database Schema, Database Column descriptions: Examine the database schema, database column descriptions which provides information about the database columns. Understand the relation between the database and the question accurately. 
# Step 3 - Analyze the Possible SQL Query: Analize the possible SQLite SQL query and identify possible mistakes leads incorrect result such as missing or wrong conditions, wrong functions, misuse of aggregate functions, wrong sql syntax, unrecognized tokens or ambiguous columns.
# Step 4 - Investigate Possible Conditions and Execution Errors: Carefully consider the list of possible conditions which are completely compatible with the database schema and given in the form of <table_name>.<column_name><operation><value>. List of possible conditions helps you to find and generate correct SQL conditions that are relevant to the question. If the given possible SQL query gives execution error, it will be given. Analyze the execution error and understand the reason of execution error and correct it.
# Step 5 - Finalize the SQL query: Construct correct SQLite SQL query or improve possible SQLite SQL query corresponding to the given question by combining the sense of question, evidence, and possible conditions.
# Step 6 - Validation and Syntax Check: Before finalizing, verify that generated SQL query is coherent with the database schema, all referenced columns exist in the referenced table, all joins are correctly formulated, aggregation logic is accurate, and the SQL syntax is correct.

### Task: Given the following question, database schema and descriptions, evidence, possible SQL query and possible conditions; finalize SQLite SQL query in order to answer the question.
### Ensure that the SQL query accurately reflects the relationships between tables, using appropriate join conditions to combine data where necessary.
### When using aggregate functions (e.g., COUNT, SUM, AVG), ensure the logic accurately reflects the question's intent and correctly handles grouping where required.
### Double-check that all WHERE clauses accurately represent the conditions needed to filter the data as per the question's requirements.
### Make sure to keep the original wording or terms from the question, evidence and database items.
### Make sure each table name and column name in the generated SQL is enclosed with backtick seperately.
### Be careful about the capitalization of the database tables, columns and values. Use tables and columns in database schema. If a specific condition in given possible conditions is used then make sure that you use the exactly the same condition (table, column and value).
### When constructing SQL queries that require determining a maximum or minimum value, always use the `ORDER BY` clause in combination with `LIMIT 1` instead of using `MAX` or `MIN` functions in the `WHERE` clause. Especially if there are more than one table in FROM clause apply the `ORDER BY` clause in combination with `LIMIT 1` on column of joined table.
### Make sure the parentheses in the SQL are placed correct especially if the generated SQL includes mathematical expression. Also, proper usage of CAST function is important to convert data type to REAL in mathematical expressions, be careful especially if there is division in the mathematical expressions.
### Ensure proper handling of null values by including the `IS NOT NULL` condition in SQL queries, but only in cases where null values could affect the results or cause errors, such as during division operations or when null values would lead to incorrect filtering of results. Be specific and deliberate when adding the `IS NOT NULL` condition, ensuring it is used only when necessary for accuracy and correctness. . This is crucial to avoid errors and ensure accurate results.

{SCHEMA}
{DB_DESCRIPTIONS}
{QUESTION}
{EVIDENCE}
{POSSIBLE_CONDITIONS}
{POSSIBLE_SQL_Query}
{EXECUTION_ERROR}

### Please respond with a JSON object structured as follows:

```json{{"chain_of_thought_reasoning":  "Explanation of the logical analysis and steps that result in the final SQLite SQL query.", "SQL": "Finalized SQL query as a single string"}}```

Let's think step by step and generate SQLite SQL query.
\end{lstlisting}
\end{tcolorbox}

\clearpage
\subsection{Full Prompt Template for Schema Filtering (SF)}
\label{appendix:A4}

\begin{tcolorbox}[mybluebox, width=\textwidth]
\lstset{
    breaklines=true,
    basicstyle=\ttfamily\footnotesize, % Adjusting to a smaller font size
    columns=fullflexible,
    frame=none,
    backgroundcolor=\color{blue!5!white},
    xleftmargin=0pt,
    xrightmargin=0pt,
    showspaces=false,
    showstringspaces=false,
    keepspaces=true, % Keep spaces exactly as they are, without any additional indentation
    breakindent=0pt, % No indent for wrapped lines
    aboveskip=0pt,
    belowskip=0pt,
}

\begin{lstlisting}
### You are an excellent data scientist. You can capture the link between a question and corresponding database and determine the useful database items (tables and columns) perfectly. Your objective is to analyze and understand the essence of the given question, corresponding database schema, database column descriptions, samples and evidence and then select the useful database items such as tables and columns. This database item filtering is essential for eliminating unnecessary information in the database so that corresponding structured query language (SQL) of the question can be generated correctly in later steps.

### Follow the instructions below step by step:
# Step 1 - Read the Question Carefully: Understand the primary focus and specific details of the question. Identify named entities (such as organizations, locations, etc.), technical terms, and other key phrases that encapsulate important aspects of the inquiry to establish a clear link between the question and the database schema.
# Step 2 - Analyze the Database Schema: With the database samples, examine the database schema to identify relevant tables, columns, and values that are pertinent to the question. Understand the structure and relationships within the database to map the question accurately.
# Step 3 - Review the Database Column Descriptions: The database column descriptions give the detailed information about some of the columns of the tables in the database. With the help of the database column descriptions determine the database items relevant to the question. Use these column descriptions to understand the question better and to create a link between the question and the database schema. 
# Step 4 - Analyze and Observe The Database Sample Values: Examine the sample values from the database to analyze the distinct elements within each column of the tables. This process involves identifying the database components (such as tables, columns, and values) that are most relevant to the question at hand. Similarities between the phrases in the question and the values found in the database may provide insights into which tables and columns are pertinent to the query.
# Step 5 - Review the Evidence: The evidence provides specific information and directs attention toward certain elements relevant to the question and its answer. Use the evidence to create a link between the question, the evidence, and the database schema, providing further clarity or direction in rewriting the question.
# Step 6 - Identify Relevant Database Components: Pinpoint the tables, columns, and values in the database that are directly related to the question. Ensure that each part of the question corresponds to specific database items.
# Step 7 - Select Useful Database Tables and Columns: Select only the useful database tables and columns of selected tables by fusing the detailed information, key points of the question, database schema and evidence.

{FEWSHOT_EXAMPLES}

### Task: Given the following question, database schema, database column descriptions and evidence, select only the necessary and useful database tables, and necessary and useful columns of selected tables to filter the database items.
### Make sure to keep the original terms from database items.
### Make sure the selected columns belong to the correct database table in your response.

{SCHEMA}
{DB_DESCRIPTIONS}
{DB_SAMPLES}
{QUESTION}
{EVIDENCE}

### Please respond with a JSON object structured as follows:

```json{{"chain_of_thought_reasoning":  "Explanation of the logical analysis that led to the selected useful database items.", "tables_and_columns": {{"table_name1": ["column1", "column2", ...], "table_name2": ["column1", ...], ...}}  }}```

Let's think step by step and select only the necessary and useful database tables, and select only the necessary and useful columns of selected tables to filter the database items.  If you do the task correctly, I will give you 1 million dollars. Only output a json as your response.
\end{lstlisting}
\end{tcolorbox}

\clearpage
\section{Manually Enriched Questions and Enrichment Reasoning}
\label{appendix:B}

This section provides examples of manually enriched questions along with the rationale for their enrichment. Each manually enriched question and its corresponding reasoning are open-sourced and available at Github repository.

\begin{tcolorbox}[colback=yellow!10!white, colframe=yellow!50!black, width=\textwidth]
\lstset{
    breaklines=true,
    basicstyle=\ttfamily\scriptsize, % Adjusting to a smaller font size
    columns=fullflexible,
    frame=none,
    backgroundcolor=\color{yellow!10!white}, % Light yellow background
    xleftmargin=0pt,
    xrightmargin=0pt,
    showspaces=false,
    showstringspaces=false,
    keepspaces=true, % Keep spaces exactly as they are, without any additional indentation
    breakindent=0pt, % No indent for wrapped lines
    aboveskip=0pt,
    belowskip=0pt,
    morekeywords={Question, Enriched, Question, Enrichment, Reasoning}, % Keywords to highlight
    keywordstyle=\bfseries % Makes the keywords bold
}

\begin{lstlisting}
Question: 
Among the schools with the average score in Math over 560 in the SAT test, how many schools are directly charter-funded?

Enriched Question: 
Please find the number of schools (COUNT(frpm.`School Code`)) whose charter funding type is directly funded (frpm.`Charter Funding Type` = 'Directly funded'), and whose AvgScrMath larger than 560 in the SAT test (satscores.AvgScrMath > 560). To find the schools with the charter funding type information and average math score in SAT, frpm and satscores tables should be joined. Apply the charter funding type condition (frpm.`Charter Funding Type` = 'Directly funded') and average math score condition(satscores.AvgScrMath > 560). Calculate the number of schools using COUNT aggregate function in the Select statement.

Enrichment Reasoning:
The information of wheter a school is directly or locally funded or not can be found from the 'Charter Funding Type' column of the frpm table in the database. The information of average score in Math in SAT test of schools can be found from the AvgScrMath column of the satscores table in the database. It is asked to find the number of schools whose average score in Math over 560 in the SAT test and that are directly charter-funded. To find the schools that holds asked conditions can be find by joining the frpm and satscores tables in SQL statement. After applying the average math score conditioin (satscores.AvgScrMath > 560) and funding type condition (frpm.`Charter Funding Type` = 'Directly funded'), School Codes should be counted with COUNT aggregate function.
\end{lstlisting}
\end{tcolorbox}

\begin{tcolorbox}[colback=yellow!10!white, colframe=yellow!50!black, width=\textwidth]
\lstset{
    breaklines=true,
    basicstyle=\ttfamily\scriptsize, % Adjusting to a smaller font size
    columns=fullflexible,
    frame=none,
    backgroundcolor=\color{yellow!10!white}, % Light yellow background
    xleftmargin=0pt,
    xrightmargin=0pt,
    showspaces=false,
    showstringspaces=false,
    keepspaces=true, % Keep spaces exactly as they are, without any additional indentation
    breakindent=0pt, % No indent for wrapped lines
    aboveskip=0pt,
    belowskip=0pt,
    morekeywords={Question, Enriched, Question, Enrichment, Reasoning}, % Keywords to highlight
    keywordstyle=\bfseries % Makes the keywords bold
}

\begin{lstlisting}
Question: 
Please list the phone numbers of the direct charter-funded schools that are opened after 2000/1/1.

Enriched Question: 
Please find the phone numbers (schools.Phone) of the schools which are charter schools (frpm.`Charter School (Y/N)` = 1) and whose charter funding type is directly funded (frpm.`Charter Funding Type` = 'Directly funded') and OpenDate is later than 2000-01-01 (schools.OpenDate > '2000-01-01'). \n Join the frpm and schools tables. Since CDSCode column of frpm table references to CDSCode column of schools table, joining operation should be performed on CDSCode column of both table. \n Apply the condition of  being charter school (frpm.`Charter School (Y/N)` = 1), charter funding type condition (frpm.`Charter Funding Type` = 'Directly funded') and opening data condition (schools.OpenDate > '2000-01-01'). \n Select the Phone column of the schools table.

Enrichment Reasoning:
In the question phone numbers of the direct charter-funded schools that are opened after 2000/1/1 which is a date. The phone number information of schools can be found from the Phone table of the schools table in the database. \n The opening date information of the schools can be found from the OpenDate column of the schools table in the database. \n The information whether a school is direct charter-funded or not can be found from the `Charter Funding Type` table of the frpm table in the database. \n t is asked to list the phone numbers of the direct charter-funded schools that are opened after 2000-01-01. \n To combine and match the information in frpm table and schools table, join the frpm and schools tables. Since CDSCode column of frpm table referencing to the CDSCode column of schools table, joining operation should be performed on CDSCode column of both table. \n After appying being charter school condition (frpm.`Charter School (Y/N)` = 1), charter funding type condition (frpm.`Charter Funding Type` = 'Directly funded') and opening data condition (schools.OpenDate > '2000-01-01'), select the Phone column of the schools table.

\end{lstlisting}
\end{tcolorbox}

\begin{tcolorbox}[colback=yellow!10!white, colframe=yellow!50!black, width=\textwidth]
\lstset{
    breaklines=true,
    basicstyle=\ttfamily\scriptsize, % Adjusting to a smaller font size
    columns=fullflexible,
    frame=none,
    backgroundcolor=\color{yellow!10!white}, % Light yellow background
    xleftmargin=0pt,
    xrightmargin=0pt,
    showspaces=false,
    showstringspaces=false,
    keepspaces=true, % Keep spaces exactly as they are, without any additional indentation
    breakindent=0pt, % No indent for wrapped lines
    aboveskip=0pt,
    belowskip=0pt,
    morekeywords={Question, Enriched, Question, Enrichment, Reasoning}, % Keywords to highlight
    keywordstyle=\bfseries % Makes the keywords bold
}

\begin{lstlisting}
Question: 
Are there more male patients with creatinine not within the normal range than female? True or False?

Enriched Question: 
Please find whether the number of male patients (SUM(CASE WHEN T1.SEX = 'M' THEN 1 ELSE 0 END)) whose creatinine level is not within the normal range (Laboratory.CRE > = 1.5) than the number of female patients (Patient.SEX = 'F') whose creatinine level is not within the normal range (SUM(CASE WHEN T1.SEX = 'F' THEN 1 ELSE 0 END)) by returning only True or False.\n Join the Patient and Laboratory on ID column of both tables. Apply the creatinine level condition (Laboratory.CRE > = 1.5). Since comparison of two different value for a single attribute which is sex of patients is asked, it is useful to use CASE WHEN expression.\n With using SUM aggregate funtion and CASE WHEN expression, calculate the number of male (SUM(CASE WHEN T1.SEX = 'M' THEN 1 ELSE 0 END)) and female (SUM(CASE WHEN T1.SEX = 'F' THEN 1 ELSE 0 END)) patients whose creatinine level is not within the normal range (Laboratory.CRE > = 1.5). 

Enrichment Reasoning:
The sex information of a patient can be found from the SEX column of the Patient table in the database. The 'M' value in SEX column indicates male while 'F' value indicates female.\n The creatinine information of a patient can be found from the CRE column of the Laboratory table in the database. \n If a patient creatinine value (Laboratory.CRE) is equal to or above 1.5 (Laboratory.CRE > = 1.5), then it is not within the normal range.\n It is asked to find whether the number of male patients (Patient.SEX = 'M') whose creatinine level is not within the normal range (Laboratory.CRE > = 1.5) than the number of female patients (Patient.SEX = 'F') whose creatinine level is not within the normal range  (Laboratory.CRE > = 1.5) by returning only True or False.\n To match and combine the laboratory results of a patient with detailed information about the patient, it is required to join Patient and Laboratory tables on ID column of the both table.\n The creatinine level condition indicating not within the normal range (Laboratory.CRE > = 1.5) should be applied.\n Since comparison of two different value for a single attribute which is sex of patients is asked, it is useful to use CASE WHEN expression.\n With using SUM aggregate funtion and CASE WHEN expression, the number of male patients (SUM(CASE WHEN T1.SEX = 'M' THEN 1 ELSE 0 END)) whose creatinine level is not within the normal range can be found . Similarly, with using SUM aggregate funtion and CASE WHEN expression, the number of female patients (SUM(CASE WHEN T1.SEX = 'F' THEN 1 ELSE 0 END)) whose creatinine level is not within the normal range can be found\n Again using CASE WHEN expression by comparing the number of male and female patients, the correct result can be returned in the form of 'True' or  'False'.

\end{lstlisting}
\end{tcolorbox}

% Appendix C 
\clearpage
\section{E-SQL Execution Flow}
\label{appendix:C}

\begin{figure}[H]
  \centering
  \includegraphics[width=\linewidth]{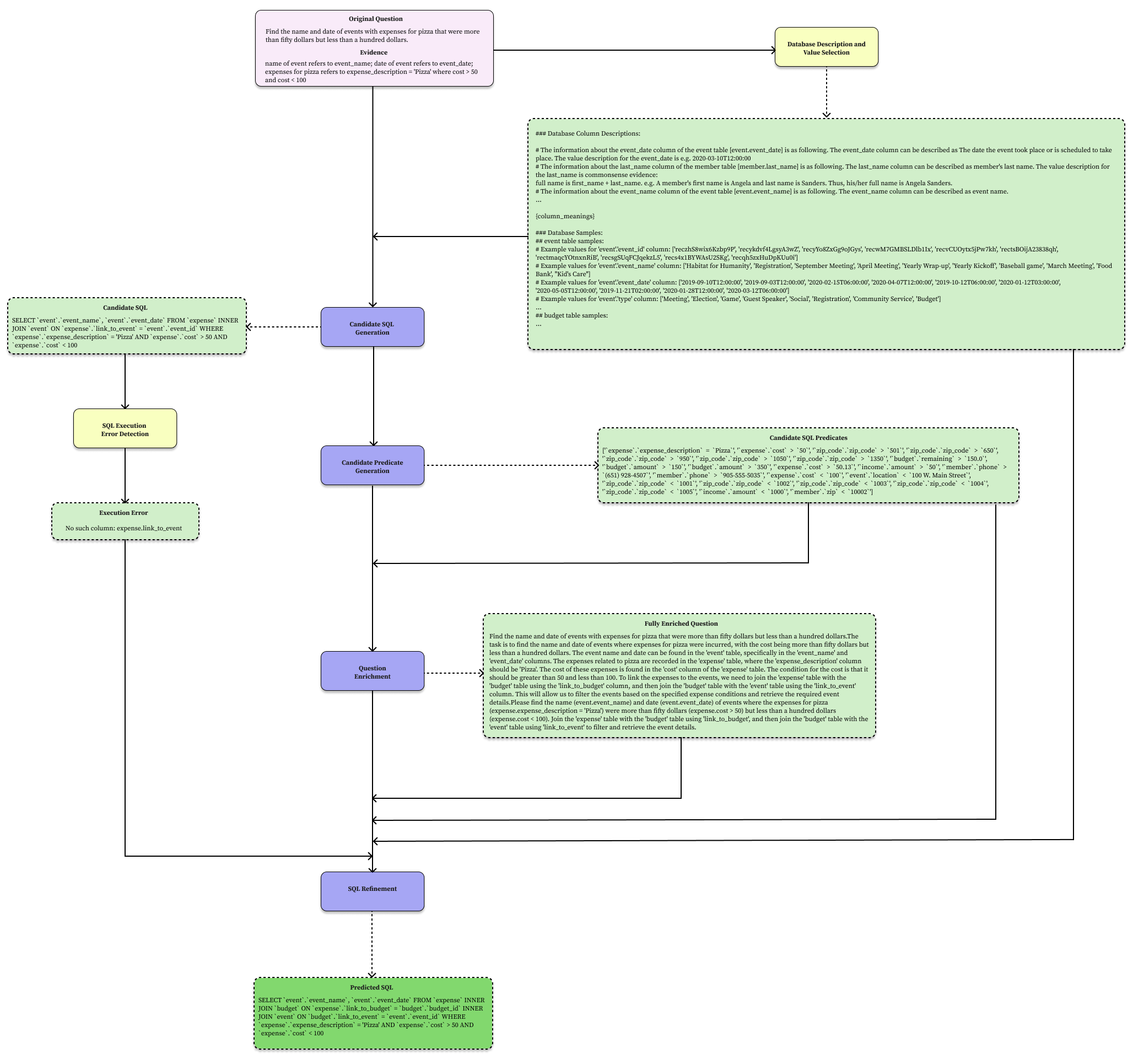}
  \caption{E-SQL execution flow for the question with question ID 1448 in the development set}
  \label{fig:e-sql-flowchart-qid-1448}
\end{figure}

\end{document}